\documentclass[11pt,a4paper]{article}
\usepackage[a4paper, total={6.5in, 10in}]{geometry}

\usepackage{hyperref}
\usepackage{xcolor}
\definecolor{darkblue}{rgb}{0, 0, 0.5}
\hypersetup{colorlinks=true,citecolor=darkblue, linkcolor=darkblue, urlcolor=darkblue}

\usepackage[round]{natbib} 
\usepackage{booktabs} 
\usepackage{courier} 
\usepackage{amsmath}
\usepackage{amsfonts}
\usepackage{amsthm}
\usepackage{latexsym}
\usepackage{graphicx} 
\usepackage{qtree}
\usepackage{color}
\usepackage{url}
\usepackage{verbatim}
\usepackage{comment}
\usepackage{algorithmic}
 
\usepackage{graphicx}

\usepackage{anyfontsize}
 
\usepackage{float}
\usepackage{caption} 
\usepackage{subcaption}

\graphicspath{{influence_matrices}}

\newcommand{\Pc}{\mathcal{P}}

\DeclareMathOperator{\algtree}{Alg_{Tree}}
\DeclareMathOperator{\logdet}{logdet}
\DeclareMathOperator{\flatmat}{Flat}

\newcommand{\newpara}{} 

\newcommand{\Prob}{\mathbb{P}}

\newcommand{\E}{\mathbb{E}}

\newcommand{\Cov}{\mathrm{Cov}}

\newcommand{\logdetnj}{\textsc{logdet+nj\ }}

\newcommand{\namecite}[1]{\citet{#1}}

\begin{document}
 
 
\title{Syntactic structures and the general Markov models}
\author{Sitanshu Gakkhar, Matilde Marcolli}
\date{April 15, 2022}

\maketitle
 
\begin{abstract}
 
We study phylogenetic signal present in syntactic information by considering the syntactic structures data from \namecite{langelin}, \namecite{collins_sswl}, \namecite{longo2020} and \namecite{koopman_sswl}. Focusing first on the general Markov models, we explore how well the the syntactic structures data conform to the hypothesis required by these models. We do this by comparing derived phylogenetic trees against trees agreed on by the linguistics community. We then interpret the methods of \namecite{longo2020} as an infinite sites evolutionary model and compare the consistency of the data with this alternative. The ideas and methods discussed in the present paper are more generally applicable than to the specific setting of syntactic structures, and can be used in other contexts, when analyzing consistency of data with against hypothesized evolutionary models. 
 
\end{abstract}
 
 
\section{Introduction}
 
The focus of the present paper is to investigate the following questions: 
to what extent syntactic features capture phylogenetic relationships and to what extent Markov models are a viable assumption for phylogenetic reconstruction based on syntactic features. For the second, we also consider an alternative that we argue approximates the infinite site evolutionary model. These questions are motivated by the fact that at both lexical and syntactic level, Markov processes are commonly assumed to underlie
computational models of language change; for instance, within the Principles and Parameters setting relevant here, \namecite{berwickniyogi97} developed models of language acquisition and language change based on a Markov process in a space of syntactic parameters. In this paper we focus only on language change processes, viewed through the lens of phylogenetic trees of language families. While the model we consider are not directly related to models of language acquisition and parameter setting, the historical changes of syntax within and across language families, through the modification of syntactic parameters, can be seen as an effect of such underlying dynamics. 
It is therefore important to develop specific methods and measures to test the extent to which a Markov model assumption is valid. Such methods will also provide tests for the presence of homologous traits, as we will discuss.

We are specifically interested in investigating the adequacy of Markov processes
in modelling phylogenetic trees of language families, based on data of syntactic ``parameters" 
(or ``features": binarized present/absent syntactic characters). We will, in particular, also
show that syntactic data do not appear to be reliable in reconstructing remote phylogenetic 
relationships. 
 
This complements other recent work towards understanding the extent to which linguistic relationships can be reconstructed based on syntax, discussed with a Bayesian inference approach in \namecite{ceolin2021boundaries,longo2020}. 
Typically, mathematical methods for phylogenetic trees reconstruction are based on an underlying assumption that the stochastic variables involved (in our case syntactic parameters/features) would be ruled by the dynamics described by a Markov model on a tree. These Markov models on trees usually come in the form of Barry--Hartigan models, where one assumes that features evolve following a Markov process across each edge and that at each site the data patterns are independently and identically distributed.  While this hypothesis has generally been justified in the main applications to mathematical biology, some of the limitations in that setting are also understood (see \namecite{Zou} as an example of such discussions). 
The question of its direct applicability to syntactic features is more subtle. Using available databases of syntactic
structures of world's languages (which at present consist largely of Indo-European languages), 
we test the validity of this Markov model hypothesis. Additionally, with phylogenetic reconstruction increasingly relying on complex models/computational approaches that involve large number of free parameters, it becomes hard to ascertain how likely the evolutionary process described by the model is to actually occur. Testing how well the evolutionary model describes the data, un-augmented with the extra parameters that introduce over-fitting concerns is an especially important sanity check.  
 
Our main conclusion is that working with Markov models on trees provides reasonably accurate results for collections of languages within a given language family, while when the size of the tree
grows to include the simultaneous presence of different language families the tree reconstruction
becomes more unreliable.  Naturally, whenever there is weak evidence for phylogeny, tree reconstructions
are expected to become unreliable, see for example the discussion in \namecite{Pagel-Meade}. The
point here is more about the role of possible discrepancies with respect to the Markov model hypothesis.
Indeed, this is consistent with what observed using different, coding theoretic
methods in \namecite{shu_marcolli}, where it is shown that, when one includes different language
families and considers the associated lists of syntactic parameters as a binary code, one
obtains codes whose position in the space of code parameters is not compatible with a
random process of code generation of the type implicit in the usual phylogenetic Markov models.
The theoretical problem of identifying a better dynamical model, beyond the Markov processes
on trees, to describe evolution of syntactic features remains to be investigated, as well as
the relation to the models proposed in \namecite{berwickniyogi97}.
 
We consider the same data of syntactic structures from \namecite{langelin} and \namecite{collins_sswl} that were analyzed in \namecite{SOBM}, using phylogenetic algebraic geometry, which depends on a general Markov models hypothesis. Our goal is to investigate how well the model describes the evolutionary processes on natural language syntax, at the same time comparing the phylogenetic signal we obtain to that of \namecite{longo2020}. We also explore the question of metricizing the space of syntactic structures that is relevant to the persistent homology machinery used by \namecite{marcolli_topling} towards questions on the phylogenetics of language families. 
 
Additionally, we note that the methods of \namecite{longo2020} can be reinterpreted as describing an alternative which approximates the infinite sites model. We also consider this model, and point out the similarity between the results from general Markov model and the methods of \namecite{longo2020}. 
 
\namecite{longo2020} also consider Bayesian methods; indeed, Bayesian methods are common and extremely powerful. The methods we focus on are distance based for the reason that this minimized the additional assumptions one needs to make about the process; for instance, the best performing model in \namecite{longo2020} is a Gamma Site Model with a specific  substitution rate. To minimize this type of input we restrict to only what is directly computable from the data. 
 
\begin{table}[!htbp]
\begin{center}
\begin{tabular}{p{4cm}|p{1.5cm}|p{9cm}}
Model & Metric & Notes  \\
\hline
Approximate infinite sites model & modified Jaccard& Does not support back mutation: each site mutates only once along the branch. \\
\hline
General Markov model & Logdet & Supports back mutations: each edge evolves according to the associated Markov matrix. The model specializes to models reversible process when edge matrices are of form $e^{tQ}$ for a rate matrix $Q$ with $t$ representing branch length: the rate matrix is the instantaneous rate of state transition; over time the transitions accumulate. By \emph{memoryless} we will mean the case where this accumulation is through a memoryless process, i.e., the transition matrix across a branch of length $t$ is $e^{tQ}$. Each edge is allowed to have a different rate matrix. In the general case where edge Markov matrices are not generated by rate matrices, there's no intrinsic notion of distance.   \\ 
\hline
\end{tabular}\caption{Summary of the two evolutionary models considered. Both models assume that the underlying topology is a binary tree and the data at the leaves comes from identical and independent evolution\label{ev_model_summary}} 
\end{center}
\end{table}

Summarizing, there are two main closely related questions that need to be disentangled in this analysis. One is the
question of the reliability of syntactic data alone in performing phylogenetic reconstruction of trees of languages and the other is the reliability of the hypothesis that syntactic features are governed by an evolutionary process describable
as a Markov model on a tree which we introduce in the next section (and alternatively the approximate infinite sites model). The way to proceed in the analysis, so as to separate and analyze
these questions, consists of the following steps:
\begin{enumerate}
\item We use available datasets of binary syntactic features (discussed in Section~\ref{SecSyntax}) 
together with the $\logdetnj$ method (described in detail in Section~\ref{SecMath}), 
which is known to perform well on data that are generated by a Markov model on a tree.
\item With this method we generate from the data a candidate tree. This is done on different groups of languages (smaller
subfamilies, groups with weaker evidence of relatedness, etc.) to control for known effects of how the reliability of
syntactic information decreases for more remote relationships. 
\item The trees generated with this method are then compared with two other classes of trees: either those
obtained from {\em the same} syntactic data but with different methods, or trees obtained with other non-syntactic
linguistic information (lexical and morphological). 
\end{enumerate}
Comparison with trees obtained from the same data helps understanding the limitations of the
Markov model hypothesis for syntactic features, while comparison with trees obtained with other methods helps identifying the reliability of working with syntactic information alone.

\section{Preliminaries}\label{SecMath}
 
A phylogenetic tree for a set of species $X$ is a tree $T$ with an identification, $\phi:X\to \text{leaves}[T]$, of the tree leaves with the elements of the species set $X$. The root $\rho$ of the tree is a choice of a vertex of $T$. Given a rooted tree $T$ on vertex set $V$ and edge set $E$, with a partial ordering on the vertices given by distance from root, a Markov process on $T$ with state set $C$ is a family of random variables $\{\zeta_v:v\in V\}$, such that if $(u, v)\in E$ then 
$$
\Prob(\zeta_v = \alpha|\wedge_{w<v} \zeta_w) = \Prob(\zeta_v=\alpha|\zeta_u)\, ,
$$
where the $w$ with $w<v$ are all the vertices preceding $v$ in the rooted tree $T$.
 
Such a Markov process can be thought of as obtained by assigning a Markov transition matrix to each edge, governing the dynamics across it. More formally, the \emph{$\kappa$-state general Markov model on a phylogenetic tree} consists of a probability distribution over the state set assigned to the root vertex, together with an assignment of a $\kappa\times\kappa$ transition matrix to each edge. The $\kappa$-state random variable $\chi$, called a \emph{character}, evolves from the root to each leaf based on the transition matrices on the path downwards. The probability distribution of the $\kappa$ states at the leaves can be thought of as a tensor, $\Pc_T$, indexed by the possible patterns. This means that the components of the tensor are the probabilities $p_{x_1x_2...x_n}$ (with $x_i\in [\kappa]$) of the character $\chi$ having state $x_i$ at the leaf $i$ for a $n$-leaf tree. The central problem is inferring the phylogenetic $n$-leaf tree given $n$ sequences of length $t$, where by sequences we mean samples of the values the character takes at each of the leaves.
 
The inference in the Markov model is usually performed assuming that each parameter (often referred to as site in phylogenetic literature) is evolving identically and independently. While it can reasonably be assumed that the topology of the tree is identical for the evolution of each site, assuming that tree parameters are identical can be problematic; selection pressures often induce sites to evolve differently, and the location of the site in the sequence may carry meaning -- specifically here each site is a different syntactic parameter and there is no a priori reason why they should be independent or evolve identically. 
 
Following \namecite{AllRho}, for a $n$-leaf binary tree with $|E| = 2n-3$ edges, the parameter space $S$ for the $\kappa$-state Markov model sits inside $[0, 1]^N$ with $N=\kappa-1 + |E|\kappa(\kappa-1)$, and there is a polynomial map $\phi_r:S\to [0, 1]^{\kappa^n}$ which gives the joint distribution of states at the leaves. \namecite{AllRho} show that there exist polynomials, called the \emph{phylogenetic invariants}, dependent only on the tree topology, which vanish on the image of $S$ under $\phi_r$. This implies that, regardless of the exact parameter values, if the data is indeed coming from a Markov model, then it is straightforward to test if the tree topology is supported by the data. With real data the invariants do not exactly vanish, as the probabilities are not exact but only estimates; instead, the magnitude of the invariants is often used as a heuristic to select between tree topologies. Phylogenetic algebraic geometry studies the map $\phi_r$ and the phylogenetic variants. In general, statistical models where such polynomial maps from parameters space to observation space are defined can be studied through an algebraic geometry approach. The general Markov model assumes that the sites are evolving identically and independently. There are modifications that allow other possibilities, but they require a priori knowledge. There are two assumptions that are implicit in the setup of these phylogenetic models: \begin{enumerate}
\item The sites of the sequences (i.e., the samples of character values at leaves) are independent and identically distributed. 
\item The interactions between the taxa at the leaves are described by a tree. 
\end{enumerate}
 
The presence of selection pressures and hybridization, which are both common phenomena, are at odds with these two assumptions. The failure of the first can be thought of as sampling from a mixture of models where the evolution of the sites is identical, and this can be particularly confounding because of the phenomenon of phylogenetic mimicry: it is well known in the phylogenetics literature that a mixture of data from different phylogenetic models can mimic another model, in the sense that leaf pattern frequencies can match pattern frequencies that are not distinguishable (see \namecite{matsen_mimicry, vstefanko_mimicry_a,vstefanko_mimicry_b}). While \namecite{allman2012mixture} show that this mimicking is unlikely with mixtures of small number models, when the state space is not small this phenomenon is an issue for $2$-state models.
 
We try to quantify the agreement (or the violation) of these two assumptions with the data of syntactic structures. In the next sections we introduce the key ingredients of our approach: the \emph{logdet} transform and the flattenings of phylogenetic tensors. 
 
The discussion is specialized to the setting of binary characters, both because the phylogenetic algebraic geometry of the $2$-state model is the most well developed, and because this is the setting that directly applies to the syntactic structures datasets. However, the techniques carry over to characters with finite state sets; the ideas developed are more generally applicable than to the syntactic structures setting and can be used when working with such models.  
 
\subsection{Logdet transform and neighbor joining}
 
An important class of metrics relevant to our setting are \emph{tree metrics}: these are metrics on the space of leaves for which there exists a tree with edge lengths such that the distances between leaves correspond to lengths of paths on the tree. The main reason why these metrics are especially useful is that tree reconstruction algorithms that use similarity measures between the data at the leaves for constructing the tree are often well behaved for tree metrics. For example, neighbor joining \citep{saitou1987neighbor} is a commonly used tree construction method that reconstructs the correct tree topology given an $n$-point distance matrix from a tree metric on the tree $T$. 
 
For neighbor joining, the requirement of being a tree metric can be relaxed so that the reconstruction is still correct as long as each entry of the distance matrix is no more than half the minimum edge length of $T$ from the tree metric associated to $T$ (see, for instance,  \citet[Theorem~5.8]{warnow}, \citet[Theorem~7.7.5]{semple_phylo}). This means the accuracy of construction is compromised if the pairwise dissimilarity between the taxa has a large variation, especially if one assumes that larger dissimilarity corresponds to larger evolutionary distances and larger noise.   
 
\namecite{nj_revealed} note that neighbor joining greedily optimizes a global criterion -- the weighted tree length computed using Pauplin's formula. This has the effect that two most similar taxa are not necessarily guaranteed to be placed together as that may not be optimal on the full tree and adding or removing taxa to the set being considered can change relationships inferred between the remaining taxa. Additionally, if the dissimilarity estimates between a small number of taxa are noisier than the rest, on adding more taxa we expect the tree construction to improve because now the outlier has less impact on the tree length. We also empirically observe this in the datasets we consider.

In the context of the Markov model, the \emph{paralinear} distance of \namecite{lake1994}, also known as logdet transform, gives a natural tree metric. \namecite{lake1994} defines the paralinear distance\footnote{The slight issue with negative determinants in Lake's definition can be sidestepped using a constant scaling of the metric and moving it inside the logarithm.} $d(S_i, S_j)$ for two sequences $S_i, S_j$ over an alphabet $\{a_k:k\in[n] \}$ as  
$$ d(S_i, S_j) = -\log {\dfrac{\det J^{ij}}{\sqrt{\det{D^1}\det{D^2}}}} $$
where $J^{ij}$ is an $n\times n $ matrix, with $(p,q)$ entry given by the number of instances of $(a_p, a_q)$ in the sequence pair $(S_i, S_j)$, and $D^1, D^2$ are diagonal matrices with sum of rows and columns of $J^{ij}$, respectively, on the diagonal. Because the normalized frequencies approach probabilities, under the assumption that each site of the sequence is independent, this measures how far the joint distribution is from being the product of marginals. 
 
Chapter~8, Section~8.12 of \namecite{semple_phylo} gives a different formulation that is also useful. For a phylogenetic Markov model on a tree $T$, with a character $\chi$ with state set $X$ evolving on it, and for leaves $x,y$, define the matrix $J^{xy}_{\alpha,\beta} = \Prob(\chi_x = \alpha \wedge \chi_y=\beta), \alpha,\beta\in C$, and the matrix $P^{xy}$ with $P^{xy}_{\alpha\beta}= \Prob(\chi_y=\beta|\chi_x=\alpha) $. One has $P^{xy} = (J^{xx})^{-1}J^{xy}$, implying that the paralinear distance
$$d(x, y) = -\log{\dfrac{|\det J^{xy}|}{\sqrt{\det J^{xx}\det{J^{yy}}}}}$$ 
becomes
$$d(x, y) = -\dfrac{1}{2}\log{\det{P^{xy}P^{yx}} }. $$
From this observation it is easy to see that, if $S_1, S_2$ are independent sequences, then $d(S_1, S_2) = \infty$, since $P^{xy}$ has rank $1$. 
 
\citet*[page~212]{fels2004} notes that the logdet transform fails to be additive when the stationary distributions for the edge transition matrices do not agree. More generally, it fails when the distribution at the root is not uniform. 
The assumption of a uniform distribution at the root is not very realistic for the syntactic data considered
here. We note that the phylogenetic algebraic geometry analysis of syntactic structures in \namecite{SOBM} 
does not require uniform distribution at the root. In the setting we consider here the imbalance between the two states is not large, and we assume neighbor joining is tolerant of this deviation from the assumption of uniform distribution at root.
 
With this last caveat, we have a natural tree metric on the space of sequences coming from a general Markov model; combining it with neighbor joining, the tree topology can be recovered. We will work with the tree constructed using logdet transform and neighbor joining throughout the next sections, and we refer to the tree constructed like this as the \logdetnj tree.  
 
\subsection{Logdet transform and rate matrices}
 
An important special case is when the edge transition matrices have form $e^{Qt}$ for a real parameter $t$ and a matrix $Q$ called the \emph{rate matrix}.\footnote{Rate matrix is any matrix where each row sums to zero, and all entries are positive off diagonal and non-positive on it; each edge is thought of as a continuous Markov chain associated to the rate matrix.} Rate matrices characterize the instantaneous rate of transition between the states for the character and the parameter $t$ can be thought of as time. An important consideration is whether every edge transition matrix, $M_e$ can be given by a single choice of rate matrix, $M_e=e^{Qt_e}$; such a process is a \emph{stationary} process. It is \emph{reversible} when $Q$ is symmetric, the reason being that the parameter $t_e$ associated to each edge can now be interpreted to mean time. \emph{Heterotachy} is the phenomenon where more than one rate matrix is involved, which significantly complicates the analysis.
\namecite{semple_phylo}, Section 8.5, notes 
that for a stationary, reversible process, the logdet transform is closely related to the expected number 
of substitutions along the edges, which is considered as the \emph{evolutionary distance}.  
 
Consider the covariance $\Cov(C, B)$ of character values at the leaves, i.e.~Bernoulli random variables $B, C$ that evolved from the root $R$ on the tree, co-evolving till the interior node to state $I_{BC}$, then evolving independently. Assume $C = M_CI_{BC}, B=M_BI_{BC}$ for Markov transition matrices, $M_C, M_B$. By the Law of Total Covariance
$$\Cov(C, B) = \E[\Cov(C, B|I_{BC})] + \Cov(\E[C|I_{BC}], \E[B|I_{BC}]),$$
where $\E[\Cov(C, B|I_{BC}))]$ vanishes, since conditioned on $I_{BC}$, $B, C$ are independent, while the covariance $\Cov(M_CI_{BC}, M_BI_{BC})$ becomes proportional to the variance of the internal state $I_{BC}$, involving the entries of $M_C, M_B$. 
 
 
When the state space is large, or otherwise when the variance is expected to become proportional to the parameter $t$, the covariance at the leaves encodes the topology of the tree that can be recovered by a simple greedy strategy: compute all pairwise covariances between the $n$ leaves, 
group the two leaves with largest covariance, and then replace the leaves that were just grouped by the single node. The covariance between this node and the other leaves is the minimum of the covariances against the constituents of the node. The process is iterated until every leaf is absorbed into some node, the covariances between two nodes consisting of multiple leaves being computed analogously. The correctness of this follows, since if we assume the evolution happens on the tree then, up to estimation errors, the minima of the covariances between leaves in different nodes should all be the same.   
 
In particular, when appropriate assumptions (stationarity, reversibility, proportionality to the time parameter) are met, the disagreement between covariance and the \logdetnj tree indicates heterotachy. This suggests that comparison of the \logdetnj tree and the tree based on covariances can be revealing in general.
 
\subsection{Flattenings, splits and phylogenetic invariants}
 
For a tree $T$ with leaves $L_T$, a \emph{split} is a partition of the set $L_T$ that is induced by deleting an edge of the tree. A partition into sets $A, B$ of $L_T$, not necessarily a split of the leaves, associates a $\kappa^{|A|}\times \kappa^{|B|}$ matrix to the partition, called the \emph{flattening} of the probability tensor, $p_{x_1x_2...x_n}$, where we are using the sets $A=\{a_i:i\in[|A|]\}, B=\{b_j:j\in[|B|]\}$ to index the tensor:
$$[\flatmat_T(A, B)]_{s_1...s_{|A|}, t_1\dots t_{|B|}}  = p_{l_1...l_{|L_T|}}$$ where $l_k$ is either $t_j$ or $s_i$, depending on whether the leaf $k$ is $b_j$ or $a_i$. 
 
The rank of the flattening $\flatmat_T(A, B)$ is $\kappa^{\min (1+e(A, B), |A|, |B|)}$ where $e(A, B)$ is the number of edges shared by the subtrees of $T$ obtained by restricting to the leaves $A$ and $B$ (see \namecite{pachter_algstats}, Theorem 19.5, and \namecite{AllRho}, Section 4). If $A, B$ is a split, then the rank is $\kappa$, and in particular all $(\kappa+1)\times (\kappa+1)$ minors have vanishing determinants. \namecite{eriksson_thesis} gives a simple way of constructing phylogenetic trees from character data for $n$ taxa by iteratively joining pairs of taxa, so that the rank of flattening matrices between the pair and the rest of vertices is closest to $\kappa$. 
 
\namecite{AllRho}, Theorem 4, shows that for the case of binary trees, for the $2$-state general Markov model, the phylogenetic ideal is generated by the $3\times 3$ minors of the flattening matrix for splits induced by each of the edges of the tree. For examples of calculations of phylogenetic invariants for the some language families see \namecite{SOBM}.

\section{Testing consistency with Markov models}
 
The starting point is a $n\times t $ matrix of data, where each of the $n$ sequences, with $t$ sites, are from a single taxon, with a particular site across sequences representing the evolution of a single character on the tree. Specifically, we work with the data of syntactic features described in the next section -- each feature corresponding to a character. There are two major checks that are needed: checking if each site represents an independent and identically evolving copy, and if the tree topology is sufficient.
 
Note that the character evolution model interacts with the data of the leaf sequences: for instance, if one uses a
stochastic Dollo model as in \namecite{Nicholls-Gray-2008}, with a large number of unique traits, the tree will
be poorly resolved. Moreover, the issue of i.i.d.~variables in syntax is delicate since it is known that syntactic
traits are not independent. We will return to discuss this issue in \S \ref{SecSyntax}. 
 
\subsection{Maximum likelihood statistics}
 
The sufficiency of the tree topology is explored using the maximum likelihood on the \logdetnj tree along with behavior of the rank of the flattening matrices. The key idea here is that if the Markov model does indeed describe the data, then the $\logdetnj$ tree reconstruction will in the limit give the correct tree topology, $T$. Coupled with a maximum likelihood estimate of the tree parameters, this gives a way to generate an empirical null distribution against which the statistics can be tested. Generating data from the maximum likelihood model, the distribution of distances from the flattening to the nearest matrix of appropriate rank (where we know the behavior of the ranks of flattening matrices from last section) provides the expectation against which we examine the actual data. Testing identical evolution in the syntactic structures data is theoretically not possible since there is only one sample from each structure. We do have a proxy that is sufficient (though not necessary) as evidence of failure of independent evolution, simply by considering the joint distribution of two fixed sites against the product of individual site distributions. Again, the expectation for this statistic can be empirically estimated using a maximum likelihood model. The simulated data come from i.i.d. simulated trials on the maximum likelihood trees, this makes z-score usable to characterize the discrepancy of the actual versus the simulated.  
 
\subsection{The influence of sites in leaf sequences}
 
A secondary question in the syntactic parameter setting is determining if there are parameters that carry higher relevance than others to determine the relationships between languages in families. We examine this using the idea of \emph{influence} from analysis of boolean functions (see \namecite{odonnell_boolean}). The tree on $n$ leaves can be represented as
a partially ordered collection of subsets of leaves with the order induced by the tree structure. Any algorithm $\algtree$ for tree construction can be thought of as a map into the collection of all subsets of the leaves, taking a value one if that subset is present in the output tree representation and zero otherwise. Each site in each sequence in the data affects the output of $\algtree$. The influence of the variable is the probability that changing the value of that variable changes the function. 
 
Intuitively, one expects that a few sites in the data should have a small effect on the reconstruction process. Moreover, on average over the data distribution, assuming i.i.d. evolution of characters, the expectation is that all sites would have similar influences. However, given this particular data sample, and absent any statistical evidence for or against identical evolution, we can hope to get some insight by considering the homogeneity of sites influences. This can also be revealing about syntactic structures themselves, as well as what structures are key in determining relationships within language families.  
 
We flesh out these ideas more concretely after introducing syntactic structures in more detail in the next section.   
 
\section{Syntactic structures: Background} \label{SecSyntax}
 
\namecite{chomsky_govt}, along with \namecite{ChoLa}, introduced the Principles and Parameters model of syntax, hypothesizing that syntactic structures for natural human languages can be parameterized by a universal set of binary variables: each variable indicating the validity of a syntactic construction in that language. The Chomskian theory of generative linguistics is now no longer based on
this Principles and Parameters model, after the Minimalist Program became a viable alternative within the field. Consequently,
syntactic features tend to be seen more in a descriptive role, rather than as claims about Universal Grammar. 
In subsequent work, including \namecite{langelin}, \namecite{collins_sswl} and \namecite{wals}, various families of syntactic features that can be formulated in binary form have been identified and data have been collected on the values of these variables over a significant number of world's languages (although at present the Indo--European family is still much more extensively represented in the data). We consider three independent such sets: the dataset produced by the LanGeLin collaboration \cite{LongGua,LongoNew}, collecting the values of syntactic parameters based on the Modularized Global Parameterization approach developed by Longobardi, the more recent data from \namecite{longo2020} encoding nominal structures, and the database of Syntactic Structures of World's Languages (SSWL) by Koopman and collaborators.
For a recent perspective on syntactic parameters see also \namecite{Biberauer}.
 
Technically, the binary variables used in the SSWL database cannot be regarded as genuine ``syntactic parameters" in the sense of the Principles and Parameters model, because of conflation of deep and surface structures. For example, instead of a basic word order variable (BWO) as in The World Atlas of Language Structures (WALS, \namecite{wals}) Feature 81A, the
SSWL dataset has several surface word order variables such as SVO, SOV, etc. (For a discussion of
deep and surface structure in word order features see \namecite{Rizzi} and also \namecite{murawaki}.)
However, as demonstrated by previous analysis carried out on this data set (see for instance \namecite{marcolli_topling, marcolli_hk}), the SSWL data still provide valid information regarding the distribution of syntactic features across
world's languages, and historical phenomena of syntactic relatedness. The LanGeLin data can be more reliably considered as genuine syntactic parameters. For simplicity of notation, we will loosely refer to all of the syntactic features collected in these databases in the form of binary variables as ``syntactic parameters". This is partly justified by the fact that modern syntactic theory has moved toward a generalization of the notion of parameter with respect to universal grammar (UG) specific parameters, by including parameters that are constructed during language acquisition, or ``schemata" in the sense of \namecite{langelin},
where general operations are UG-specified rather than individual parameters.
For a recent general theoretical discussion of syntactic parameters, we refer the reader to \namecite{Rizzi}.
For a general introduction to syntactic structures and the parameters model, we refer the reader to \namecite{ChoLa} and to the papers collected in the recent volume, \namecite{LingAn}, which presents an up-to-date overview of the current understanding of syntactic parameters in the linguistics community.  For a non-technical introduction to syntactic parameters aimed at a general audience of non-linguists, we recommend \namecite{Baker}. 
 
To each language there is an associated vector of syntactic parameter values which gives coordinates in an ambient metric space, with the choice of metric dependent on the context. A main open question in this parametric model of syntax is identifying a good set of independent variables, or equivalently understanding relations between syntactic parameters and constraints on the locus of possible grammars inside the larger ambient space. We refer to this problem as ``the geometry of syntax". Considerable work has been done towards understanding linguistic relationships and how syntax is constrained based on this metric space structure. The latter is an interesting question from the perspective of language acquisition: within this model of syntax, it is assumed that the values of the parameters are learned in the process of language acquisition, based on exposure to a set of positive examples;  \namecite{niyogi} gives an overview of mathematical models of language acquisition within this syntactic parameter model. \newpara

\subsection{Syntactic parameters and phylogenetics}
 
Longobardi and the LanGeLin collaboration introduced the use of syntactic parameters to reconstruct phylogenetic trees of language families of interest to historical linguistics, \namecite{LongGua}. Linguistic phylogenetic trees based on data of syntactic structures were also analyzed using phylogenetic algebraic geometry in \namecite{SOBM}. Topological data analysis of syntactic structures was used in \namecite{port2018} and \namecite{marcolli_topling} to identify
historical linguistic phenomena not captured by tree structures. \newpara
 
However, as discussed in \namecite{SAHWM} and in \namecite{marcolli_topling}, prior work addressing linguistic relationships based on the analysis of syntactic structures shows certain divergences in the structure of phylogenetic trees, with respect to what is known from historical linguistics. In particular, while the phylogenetic algebraic geometry method of \namecite{SOBM} correctly selects the historically accurate tree among a preselected list of candidates, for languages belonging to preselected and sufficiently small families, tree reconstruction methods based on the use of $\ell_p$ metrics and neighbor joining, or direct application of phylogenetic packages like PHYLIP to the SSWL data, as well as trees derived from persistent components in the persistent homology computations applied to either SSWL or LanGeLin data, show some significant amount of misplacement of languages both within and across language subfamilies.\newpara
 
In the linguistic context one does not reasonably expect that all leaves are at the same distance from the root (this is known as the ``clock assumption'' in phylogenetics, see e.g. \namecite{warnow}); for example, in the family of early European languages we consider, Tocharian and Hittite are not contemporaneous with others like Albanian and Greek. It is known in the literature that metric space methods are susceptible to failure in absence of the ``clock assumption''. To see intuitively why this failure happens, notice that, when we compute distances between taxa that are far apart in time, distances measured by Euclidean metrics only see where the vectors describing the taxa disagree, and miss the differences that arose and were undone during the intervening time. This under-estimation of evolutionary distances when back mutation is allowed by the frequently used Euclidean metrics, due to missing unobserved changes in syntactic structures, approximates an unreasonable model for linguistic evolution: if such metrics are an accurate representation of the metric relationships, then that implies that a syntactic parameter flips at most once in the evolution process. This makes convergent evolution much less likely and is at odds with known historical phenomena of multiple reversals in some syntactic parameters. Further, in language evolution we do see homoplasy phenomena and horizontal transmission in syntax, as discussed for instance in \namecite{homoplasy} and detected through persistent first homology computation in \namecite{marcolli_topling} and \namecite{port2018}. 
 
This leads us to positing that the point of failure here lies in the fact that the metrics used are not capturing the evolutionary distance. The reason for the good results from phylogenetic algebraic geometry also becomes clear: the phylogenetic invariants machinery is agnostic of the metric structure and only utilizes the general Markov model derived invariants. As the logdet metric is the natural metric in the Markov model setting, we move from using invariants to select phylogenetic trees to using logdet metric to construct them. We first apply the techniques introduced to testing how well the data match the general Markov model, and then consider the constructing phylogenies using this approach.
 
\subsection{General Markov model and syntactic parameters}
 
The phylogenetic algebraic geometry methods of \namecite{AllRho}, \namecite{PaSturm} appear very successful when applied, as in \namecite{SOBM}, to trees of language families, and that in itself is evidence in favor of Markov models on trees. However, there are reasons why one can expect significant discrepancies from these models when investigating phenomena of language relatedness at the syntactic level. Markov evolution is a restricted class of models for how syntax/languages may evolve and one does not necessarily expect the relationship between any two languages to be well described by such a process. For example, languages evolving in close geographic proximity as in the case of the 
microvariation phenomena studied in \namecite{microvar}, represented in the data of Romance and Hellenic Southern Italian dialects in the LanGeLin data, can present more interaction than permitted by tree models. Known historical linguistic phenomena involving influences across different tree subbranches are well known at the lexical level (the Anglo-Norman bridge for example) but rarer at the syntactic level, although such structures are visible in the persistent first homology studied in \namecite{marcolli_topling} and \namecite{port2018} (the Gothic-Slavic-Hellenic loop discussed in \namecite{marcolli_topling} for example). Such phenomena are beyond what is describable purely in terms of Markov models on trees. Moreover, different syntactic parameters are not independent variables: some relations are explicitly known (as discussed in \namecite{LongGua,LongoNew} for instance, and also in \namecite{evolang12}), while other relations can be detected through methods of data analysis, as in \namecite{marcolli_hk,kanerva}, or through methods of coding theory \cite{shu_marcolli,marcolli_ling_ecc}. The presence of dependencies between syntactic parameters violates the Markov models on trees hypothesis that these variables can be treated as identically distributed independent random variables. Thus, the effectiveness of the Markov evolution on trees as a model for syntactic relations between languages lingers on how large the effect of such deviations may be. 

Our purpose here is to show that, despite these possible discrepancies, the statistics of the data of syntactic structures, tested over sufficient diverse language families, are largely consistent with Markov models on trees when restricting data to within given language subfamilies. We will show that the tree reconstructions obtained by this method become significantly less reliable when the size of the tree is enlarged to include different language families, as the effect of deviations from the Markov evolution hypothesis amplifies with the size of the tree. To be more precise, what we see as the size of the language set grows encompassing different language families is that misplacement errors {\em within} the subfamilies decrease, while significant misplacements {\em across} different families occur. We see this, for example, in \S \ref{GrecoRomanceSec} with the Greco-Romance tree, where some misplacements within this subtree disappear when instead of considering only this subset of languages, we consider them within the full Indo-European tree (this subset of languages has a large sampling bias, as it contains a large number of closely related Italian dialects, considered in the microvariations study of \namecite{microvar}). 
Examples of misplacements across families can be seen, for instance, in the placement of Welsh within the Germanic tree, in the case of the full Indo-European tree of \S \ref{full_ie_family}, or the fact that the non-Indo-European Dravidian languages Tamil and Telugu are placed inside the Indo-Iranian subtree of the Indo-European tree.
This points, on the one hand, to an improved performance of the neighborhood joining within subfamilies, but at the same time to a more visible discrepancy with respect to the Markov model hypothesis when different subfamilies are simultaneously taken into consideration. 
 
Understanding when the general Markov model applies, servicing the logdet as the natural evolutionary metric, also gives insight into the Geometry of Syntax paradigm of \namecite{port2018} and  \namecite{marcolli_hk} which grapple with choice of metric when trying to understand the geometry: we note that when studying evolutionary relationships, it is the evolutionary distance that should be considered.\newpara
 
We note that these databases have been updated since the analysis of \namecite{marcolli_topling, marcolli_hk}, with the SSWL dataset especially being subject to frequent additions and updates of parameter values. This results in some minor discrepancies in values of some invariants that we compute with respect to prior results, but these do not change the main conclusions. \newpara
 
\subsection{LanGeLin dataset}
 
The LanGeLin dataset collects the values of 83 syntactic parameters based on the Modularized Global Parameterization approach developed by Longobardi, for a set of 62 languages, mostly Indo-European. A complete list of the languages and parameters included in this database is reported in \S 1.2 of \namecite{marcolli_topling}. 
 
\subsection{Entailment in the LanGeLin dataset}
 
The syntactic parameters from Longobardi's LanGeLin collaboration dataset take on values $\pm1$ as well as $0$ with zeroed values indicating dependence on other parameters. To ensure the assumption of independent evolution of parameters, we disregard all parameters that take on a zero value in the language family in consideration. When defining a metric based on these parameters, this leads to a bias towards underestimation because, when computing the dissimilarity, if the parameters underlying the dependent parameters differ then the dependent parameters will also differ. However, we have disregarded them, leading to a dampening of perceived syntactic difference. This can be viewed as a special case of not all parameters contributing uniformly to the syntax. We briefly touch on this in the discussion. This effect is also present on the SSWL dataset, though the dependence there is not explicitly identified. Note we have only removed dependent parameters that have been explicitly identified, and other dependencies may still be present in the data. 
 
In the geometry of syntax formalism, the functional dependence of zeroed parameters is exactly what defines the geometry and is of particular interest from that perspective. Since we expect this functional dependence to be different for different language families, the scheme of disregarding parameters with zeros only across the language family considered is sufficient.

\subsection{Syntactic Structures of World's Languages (SSWL) dataset}
 
The current version of SSWL dataset contains 252 languages and 115 syntactic binary variables. The list of languages and syntactic features of the SSWL dataset is discussed in detail \S 1.2 of \namecite{marcolli_topling}. The set of languages included in the database range across several non-Indo-European language families: the most represented families are, in decreasing number of languages: Indo-European, Niger-Congo, Austronesian, Afro-Asiatic.\newpara
 
An issue with the SSWL data is that the syntactic features are very unevenly mapped across the languages in the database: some languages have 100\% of the syntactic features recorded, while others are only 2\% mapped. Any subjectivity that may enter analysis in dealing with this incompleteness is removed by following the approach of previous work, where one either sets incomplete parameters to $0$ (with $\pm 1$ the binary values of recorded
parameters) or one chooses to work only with those parameters that are completely mapped for the language family under consideration (the advantages and disadvantages of these methods are discussed, for instance, in \namecite{marcolli_topling,port2018,shu_marcolli,SOBM}). Note that the second method does bias the analysis towards Indo-European languages, which tend to be more extensively mapped in the SSWL database.
 
\subsection{The \namecite{longo2020} nominal structures data} 
 
The dataset of \namecite{longo2020} encodes the nominal structures in 69 languages across 13 Eurasian families, using 94 binary variables. The dataset is significantly more complete than either the LanGeLin dataset or the SSWL dataset. The parameters show entailment like the LanGeLin dataset, with entailed parameters marked by using a zero value, as opposed to $\pm$ values otherwise. For a more complete description, we refer to \namecite{longo2020}. We note that there are sets of languages that are degenerate in this set in the sense that for all languages in these subsets all syntactic structures are identical; we only keep one representative from each subset while \namecite{longo2020} use all; we do this since keeping multiple representative adds no information but can bias neighbor-joining because of how it minimizes the balanced minimum evolution criterion, see \namecite{nj_revealed}.    
 
\subsection{Reinterpreting the \namecite{longo2020} metric} 
 \namecite{longo2020} use a modified Jaccard similarity value with Unweighted Pair Group with Arithmetic Mean (UPGMA) clustering. In their case a parameter only contributes to the syntactic distance between two languages when it is set in at least one of them:
 
$$d_{\text{modified\_jaccard}}(l_1, l_2) := \dfrac{N_{-+} + N_{+-}}{N_{-+} + N_{+-} + N_{++}}$$
 
\noindent where $N_{ab}$ is the frequency of value $a$ for parameters from language $l_1$ and $b$ for language $l_2$. This can be thought of as modelling an infinite sites evolutionary model in the sense that it is counting how many events happened in the evolution of the sequence on the tree and how many of them were different between the pair. There is no contribution from unobserved changes: in effect parameters once set are not unset till an evolutionary split happens, and along any branch site a nominal structure may change at most once. Because the number of structures that separate closely related languages is small with respect to the number of structures, this scheme approximates the infinite sites model of evolution \cite{ma_ism}. Additionally, as all languages in this dataset are currently extant, the assumption that all languages are at the same distance from the root, and therefore the choice of UPGMA reconstruction made by \namecite{longo2020} is reasonable, although this is confounded by rate variation along different branches. This approximate model is an alternative to the general Markov model, the key difference being the possibility that the same syntactic
feature could undergo multiple updates. 
 
As noted by \namecite{longo2020}, there are asymmetries  
in state transitions, with transitions primarily only observed in one direction (we see also this asymmetry in the maximum likelihood model we obtain). This asymmetry makes unobserved changes across an edge unlikely, so we expect that a highly asymmetric model will approximate this model. Pushing this a step further, if the evolutionary process is well described by such a model, then $d_{\text{modified\_jaccard}}$ would be approximately additive. We will use neighbor joining with this metric, to get the correct reconstruction guarantees that it offers, and use that to test if there is an alternative that better fits the data compared to the general Markov model. 
 
\namecite{longo2020} also present a reconstruction using Bayesian phylogenetics (built on the Markov model approach with rate matrices) obtaining results that can be considered arguably better than the UPGMA approach, giving weight to the Markov model. The point we want to make is that one does not expect the evolutionary process along any branch to be necessarily memoryless, that is, with an underlying rate matrix: the evolutionary process for syntactic parameters is less like molecular sequence evolution, which provides motivation for Bayesian phyologenetics. The closer analog in biology is stem cell differentiation, which has been modelled as a non-Markov, in the sense that the process is not memoryless -- we do not get substitutions accumulating as the exponential of a rate for the length of the branch \cite{stumpf2017}. In the linguistics setting, \namecite{greenhill_pacific} relate the linguistic diversification to population expansion, and social and geographical constraints on the population. From this perspective as well, it is reasonable to suppose that linguistic evolutionary processes have memory over larger timescales; the success of models with varying rates can be viewed as encoding this in the parameters that govern the rates. An approach that does not introduce this added complexity would offer robustness at the expense of some of the descriptive power of complicated models. This tempers the concerns about use of the extra parameters,  and provide a way to validate the more complex models. For this reason, one would like to evaluate the Gamma site model of \namecite{ceolin2021boundaries} against a rate matrix agnostic alternative.   
 
\smallskip
 
\subsection{Other linguistic data}
 
As discussed above, our analysis in this paper is based entirely on syntactic data, organized in the form of syntactic features, which are binarized (present/absent) syntactic characters. The properties of the model depend on the nature of the character data. For example, using binary as opposed to multistate characters leads to different mathematical properties of the corresponding phylogenetic model. When lexical data are used for phylogenetic analysis in linguistics, these aspects have a significant impact on the analysis and are discussed in depth. For a detailed discussion of Bayesian phylogenetic analysis based on lexical data see for instance \namecite{Greenhill}. Morphological data have also been used for phylogenetic analysis in linguistics, in a binarized form, for example in \namecite{RWT}. 
There is a significant difference in the use of syntactic data, namely the fact that syntactic features are two-state reversible, unlike lexical features, as we have discussed at length above. We will return in the following sections to point out where this needs to be taken into consideration in the analysis. 
 
\section{Markov evolution in language families}\label{sswl_longo_analysis}
 
The LanGeLin and SSWL datasets are still active projects and only partially complete, with Indo-European languages being most completely defined. So considerable prior analysis has focused on Indo-European languages. To test the ideas put forward in the previous section, we consider the following five groups of languages studied in \namecite{marcolli_hk, marcolli_topling, shu_marcolli}, which
include three sub-families of the Indo-European family, one hypothetical macro-family, and a small set of early attested
Indo-European languages, 
\begin{enumerate}
\item {\em Germanic}: Dutch, German, English, Faroese, Icelandic, Swedish.
\item {\em Slavic}: Russian, Polish, Slovenian, Serb-Croatian, Bulgarian.
\item {\em Romance}: Latin, Romanian, Italian, French, Spanish, Portuguese. 
\item {\em North Eurasian}: Finnish, Estonian, Hungarian, Khanty, Udmurt, Yukaghir, Turkish, Buryat, Yakut, Even, Evenki
\item {\em Early Indo-European}: Hittite, Tocharian, Albanian, Armenian, Greek.
\end{enumerate}

\subsection{The choice of language sets}
The last family listed above includes some of the early branchings of the Indo-European family tree. Clearly, it might have been preferable 
to select a different subset of early attested Indo-European languages, perhaps including Sanskrit, Avestan, Old Church Slavonic (OCS), etc.
Unfortunately, these are at present not included (or extremely incompletely mapped) in the available syntactic databases, so they could not be used. Currently, the only languages in this early IE group that have enough syntactic features data for any kind of comparative analysis are those listed above. However, the choice of this specific set of languages is significant for the following reason.
There has been some debate in recent years in the linguistic community (see \namecite{IEcontroversy}) around computational reconstructions of the structure of the Indo-European tree near the root. In particular, this subset of languages was chosen in \namecite{SOBM} in order to compare the relative positions of the Anatolian and Tocharian branches and the Albanian, Armenian, and Hellenic branches, between two candidate trees, one obtained in \namecite{bouckaert} on the basis of lexical data and one, generally regarded by linguists as more reliable, obtained in \namecite{RWT} including morphological data. The phylogenetic algebraic geometry method, applied in \namecite{SOBM} to the SSWL syntactic data for this set of languages, slightly favors the tree of \namecite{RWT}. While some of these ancient languages, like Ancient Greek, are very completely mapped in the SSWL database, others like Hittite and Tocharian are only very coarsely mapped. This implies that there are only 22 variables in the SSWL dataset that are fully mapped for all of these ancient languages. Since the analysis in this case is based on a very small set of syntactic features, it should be regarded as less reliable than the cases of the other families above, for which a larger set of completely mapped parameters is available.  
We use the combined set of parameters from SSWL and LanGeLin datasets when the languages in consideration are present in both databases. For Romance and Slavic families which are present in both databases we use the combined parameter values from both, restricting to parameters which are set in all languages only. So, this is the analysis that is based on the most complete set of data; although still having to drop partial unset parameters is not ideal. For the other families too, as in \namecite{SOBM}, we only use parameters that are set for all languages.  
 The results obtained in this way are discussed in the following subsections; we defer the discussion of romance family to the end, as here we find that not only is the \logdetnj tree different from historically correct tree, but also has a lower phylogenetic invariant. \newpara
 
Working in the setting of binary syntactic structures, the general Markov model setup is specialized to binary characters. The sequences at leaves are also binary and the transition matrices are $2\times 2$. One could consider the unset parameter to be a third symbol, however, the statistics when a parameter is unset in both languages under consideration become ambiguous, so working only with those parameters that are completely mapped for the selected subfamily of languages is favored. \newpara
 
We first construct $\logdetnj$ trees for these families and evaluate how consistent these are with what is accepted in the linguistics community. Note that the trees constructed are unrooted. This is the case also when one applies the phylogenetic algebraic geometry methods (this issue for linguistic phylogenetic trees is discussed in detail in \namecite{SOBM}). In particular, the placement of the root is related to the knowledge about ancient languages in the database. While for the Indo-European language
family, several ancient languages are represented in the data, and this information can be used to correctly root the trees, for language families where only the modern languages are represented in the data one can obtain the information on the tree topology but not as a rooted tree.  \newpara
 
By the arguments outlined above, under the assumption of general Markov model (including the uniform distribution at the root), the $\logdetnj$ tree will recover the correct tree. The reconstructions for Germanic, Slavic and Uralic and Altaic languages are briefly discussed before we focus on the two cases which lend themselves to a richer analysis.   
 
\subsection{Germanic languages}

The \logdetnj tree constructed for the Germanic family using the 89 completely mapped syntactic parameters correctly identifies the separation between West Germanic (Dutch, German), and the East Germanic (Swedish, (Icelandic, Faroese)).
The \logdetnj trees are unrooted.
A common method of rooting trees by choosing an outgroup representative is not meaningful here, since the outgroup element may not be evolutionarily related, or the evolutionary distance may be so large that the noise in estimating it will significantly affect the results. Thus, we have simply placed the root in the tree where it is known to be from 
historical linguistic information, while the \logdetnj tree is simply providing the tree topology.
 
\begin{figure}[!htb]
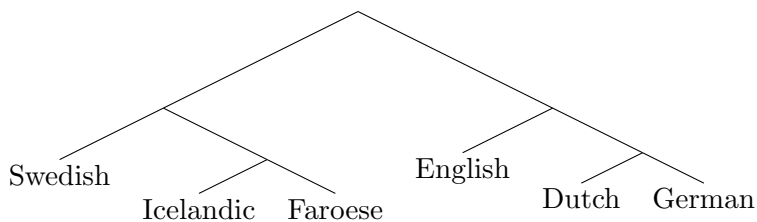

        $$ \Tree[ [ Swedish [ Icelandic Faroese ] ] [ English [ Dutch German ] ] ] $$
        \caption{The Germanic \logdetnj tree,  showing the East and West Germanic split, with the root placed 
        according to the historically accepted tree.}
\end{figure}
 
The reconstruction is very robust and with $\approx60\%$ of the data, we can recover this topology with probability approximately $0.7$.  
 
We note that using various euclidean ($\ell_p$) metrics (with UPGMA tree construction) fails to recover the East/West Germanic split, as does in the tree of the persistent connected components of \S 6.4 of \namecite{marcolli_topling} which mixes North and West Germanic languages. The logdet transform is a better proxy for evolution distance than $\ell_p$ metrics. Specifically, $\ell_p$ metrics do not account for unobserved changes in the syntactic structures. This observation underscores using evolutionary distance to explore the geometry of syntax over embedding into other metric spaces.

\subsection{Slavic languages}

For the Slavic family, there are 68 parameters that are fully mapped between the two datasets: 45 for LanGeLin and 23 for SSWL. With the most recent version of the SSWL data, these parameters do not separate Slovenian from Russian, so for this reason we have excluded Slovenian.  The previous version of the data used in the phylogenetic algebraic geometry analysis
of \namecite{SOBM} correctly placed Slovenian close to Serb-Croatian, in the South Slavic subbranch, while with the later
version of the data used in the persistent components tree of \namecite{port2018}, the current ambiguity is resolved by
(incorrectly) placing Slovenian next to Russian. \newpara

\begin{table}[!htb]
    \begin{center}
\begin{tabular}{lrrrrr}
    \toprule
    {} &   Russian &    Polish &  Slovenian &  Serb-Croatian &  Bulgarian \\
    \midrule
    Russian       &  0.000000 &  0.127036 &   0.000000 &       0.029729 &   0.092947 \\
    Polish        &  0.127036 &  0.000000 &   0.127036 &       0.160433 &   0.232805 \\
    Slovenian     &  0.000000 &  0.127036 &   0.000000 &       0.029729 &   0.092947 \\
    Serb-Croatian &  0.029729 &  0.160433 &   0.029729 &       0.000000 &   0.126210 \\
    Bulgarian     &  0.092947 &  0.232805 &   0.092947 &       0.126210 &   0.000000 \\
    \bottomrule
\end{tabular}\caption{Loget distance matrix for Slavic family including Slovenian}
\end{center}
\end{table} 

There are still 68 parameters across SSWL and Longobardi datasets that are mapped for the four remaining languages. The SSWL parameters for this family are very poorly mapped, and the distance matrix  for the SSWL data alone is highly degenerate. 
We work with the full collection of 68 parameters spanning the two datasets. Constructing the \logdetnj tree we get
\begin{figure}[H]
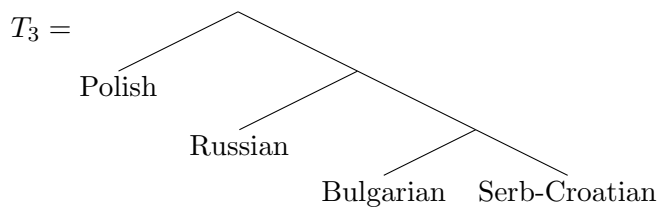

    $$T_3= \Tree [ Polish  [ Russian [ Bulgarian Serb-Croatian ]  ] ]$$
    \caption{Slavic languages tree}
\end{figure}
Here again the tree is unrooted, and we have chosen to draw it so that the root is 
placed consistent with historical linguistic knowledge. Since there are only four branches, 
the only information contained in the tree topology is the placement of the unique
internal edge, namely splitting of the leaves into two pairs of adjacent vertices, $\{ \text{Polish}, \text{Russian} \}$
and $\{ \text{Bulgarian}, \text{Serb-Croatian} \}$, 
which here correctly reflects the grouping together of the 
South Slavic branch. 
 
With Slovenian excluded, all three \cite{Longo2013,SOBM,nurbakova} agree on this tree. On the other hand,
the position of Slovenian in these trees is not consistent: both \namecite{Longo2013} and \namecite{SOBM} place it 
adjacent to Serb-Croatian and separated from the Russian and Polish subtree,  while 
\namecite{nurbakova} places Slovenian as an early branch separated from both the subtree with 
Serb-Croatian and Bulgarian and the subtree with Russian and Polish.

The reconstruction is robust: with approximately $60\%$ of the parameters sampled uniformly randomly, this topology appears with a probability in excess of $0.6$. An alternate topology which places Russian with Bulgarian also appears in some subsamplings; this proximity of Bulgarian to Russian is also observed in the tree of 
persistent connected components from \namecite{port2018}.

\subsection{North Eurasian languages}

We consider here a collection of languages encompassing the Uralic, Altaic, and Tungusic languages
available in the databases. These are languages that are not usually grouped together as a family.
In fact, the evidence for affinity is very weak (even for the Uralic group). We will see
that, as predictable in a similar situation, the $\logdetnj$ performs less well, although
it still contains valid information. 
 
As in \namecite{marcolli_topling}, we consider the languages belonging to the Uralic family 
(Estonian, Finnish, Hungarian, Udmurt, Yukaghir, Khanty) and to the more hypothetical
Altaic family (Buryat, Turkish, Yakut, including the Tungusic languages Even and Evenki).
As we discuss below, Yukaghir is usually considered an independent language which is 
likely to be related to the Uralic family. 
 
Compared to what was obtained by the topological method of \namecite{marcolli_topling}, 
for the North Eurasian $\logdetnj$ recovers a tree that appears more 
consistent with known linguistic relations. We first consider the full set of Altaic-Uralic-Tungusic languages
listed above, using 28 parameters from the LanGeLin dataset to obtain the (unrooted) tree:
 
{\scriptsize
$$\Tree[  [ [ [ Buryat Turkish ]  Yakut ]  [  [ Evenki Even ] Yukaghir  ] ]   [ Udmurt  [ Khanty  [ Hungarian  [ Estonian Finnish ] ]]]] $$
}
 
The $\logdetnj$ tree is unrooted: we have drawn it for convenience so that the root is placed at the divide between the Uralic and the Altaic-Tungusic languages, but it is important to keep in mind that this is not meant to represent a correct historical linguistic rooting of the tree, only a graphical convenience. Notice that here the two groups of languages are clearly separated, with the only misplacement, with respect to this divide, consisting of the Uralic language Yukaghir that is placed together with the Tungusic languages Even and Evenki. However, several misplacements are present within the tree. 
 
This tree recovers the large-scale structure of the family: Udmurt separates out from the rest in the Uralic subtree (Khanty, Hungarian, Finnish, Estonian), which is sensible as it is the lone representative of Permic branch of Uralic languages, while in \namecite{marcolli_topling} it was placed into the Altaic subtree. The rest of the Altaic subtree (without the Tungusic languages and the misplaced Yukaghir) appears in the form (Yakut (Buryat, Turkish)). This is not as expected, since it does not group together the two languages (Turkish and Yakut) that belong to the Turkic subgroup, but rather groups Turkish with Buryat, which belongs to the Mongolic subgroup. 
Regarding the misplacement of Yukaghir, note that this language is considered sufficiently distinct from the Uralic languages to form an Uralo-Yukaghir meta-family and the extent of the relationship between Uralic and Yukaghir is a subject of active investigation (see for instance \namecite{Piispanen_2013}). Its placement close to the Tungusic languages Even and Evenki is more consistent with geography (Even, Evenki, Yukaghir all belong to parts of Eastern Russia) than reflecting the Uralo-Yukaghir relation. 
The position of Khanty in the Uralic subtree is not as expected: Hungarian and Khanty should separate from the Balto-Finnic subtree (as they do in the topological analysis of \namecite{marcolli_topling}). This can be compared with the historically agreed tree for the Uralic family:
$$ \Tree[ Udmurt [ [ Estonian Finnish ] [ Khanty Hungarian ]  ] ] $$
 
Examining the consistency of this construction, the topology of the \logdetnj tree can be recovered in the case of Uralic with probability a half. There are three topologies that appear in the simulated data, including the correct topology.

\subsection{Early Indo-European languages}
 In this case the sparseness of data -- there are only 22 parameters that are completely set for this set of languages -- makes this dataset difficult to work with, and conclusions drawn from the analysis should not be regarded as very reliable. There are additional issues: for example, the values of one parameter each for Hittite and Tocharian have been updated. 
\footnote{In the updated SSWl Hittite has ``11 Adposition Noun Phrase" set to value $0$ and Armenian (Western Armenian) has ``Neg 01 Standard Negation is Particle that Precedes the Verb" set to value $1$.}
since the analysis of \namecite{SOBM}, and this renders the Hittite and Tochrian degenerate on the space of the parameters that are completely set. We use this dataset after rolling the update back for comparison with \namecite{SOBM}, but this does confound the results.   
 
\begin{figure}
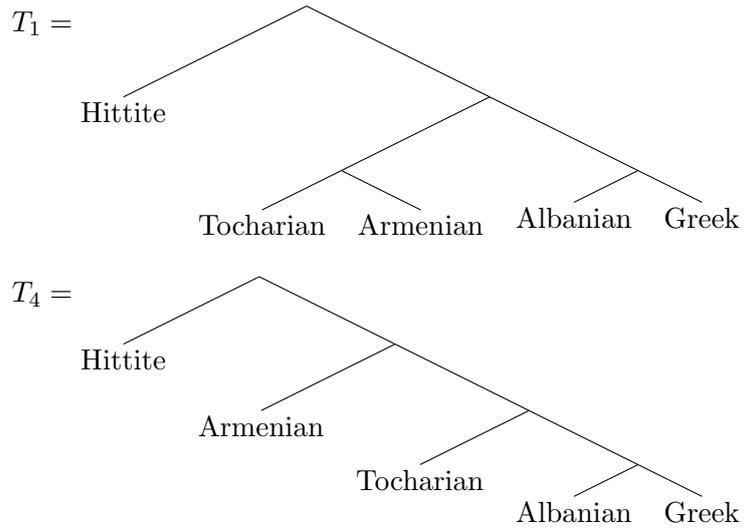

 
$$T_1=\Tree [ Hittite [ [ Tocharian Armenian ] [ Albanian Greek ]  ] ] $$
 
$$T_4=\Tree [ Hittite  [ Armenian [ Tocharian [ Albanian Greek ] ] ] ]$$ 
 
\caption{\namecite{bouckaert} tree, $T_1$, that agrees with \logdetnj tree $T_{\text{orig}}$, and the \logdetnj tree, $T_4$\label{logdet_ancient}}
\end{figure}

We note that without rolling back the two parameters, the \logdetnj tree, $T_{\text{orig}}$, for Early Indo-European languages rooted appropriately 
near the Anatolian-Tocharian split is the same as the one obtained with purely lexical data by \namecite{bouckaert}. The \logdetnj tree after rolling back the updates is $T_4$, (figure~\ref{logdet_ancient}). A different tree topology, restricted to the same subset of languages,
was obtained in \namecite{RWT}, based on a combination of lexical and morphological data. 
In \namecite{SOBM} phylogenetic invariants based on the SSWL syntactic data are computed for the two tree topologies obtained by \namecite{RWT} and by \namecite{bouckaert} and it is observed that the tree of \namecite{RWT} has a smaller resulting phylogenetic invariant. The disagreement with \logdetnj tree could be interpreted as implying that the evolutionary processes acting on syntax are again not Markov, but this is not confirmed by the phylogenetic invariants computation (also based on the Markov model) that favor the tree of \namecite{RWT}. It is possible that the discrepancy between the \logdetnj approach and the phylogenetic
approach here may reflect the fact that phylogenetic invariants, in the model of \namecite{AllRho},
allow for a nonuniform distribution at the root, while as observed earlier the logdet 
transform fails to be additive when the distribution at the root is non-uniform (\namecite{fels2004}, page~212).  Since we are looking here
at a group of languages that branched out very close to the putative root of the Indo--European tree, this issue may be significant. 
 
\begin{figure}
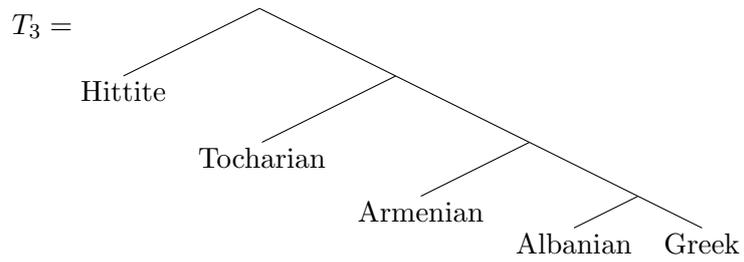

    $$T_3= \Tree[ Hittite [ Tocharian [ Armenian [ Albanian Greek ] ] ] ] $$
\caption{The tree from \namecite{rexova2003cladistic} obtained on restricting to 17 out of the 22 parameters that are recorded for this set of languages. Note that the \namecite{rexova2003cladistic} is almost the \namecite{RWT} tree, except that Armenian and Albanian are switched. \label{FigRexovaTree}
}
\end{figure}

On exploring this further we find that there exists a set of 17 parameters
\footnote{01 Subject Verb, 06 Subject Object Verb, 11 Adposition Noun Phrase, 13 Adjective Noun, 15 Numeral Noun, 17 Demonstrative Noun, 19 Possessor Noun, 21 Pronominal Possessor Noun, Neg 03 Standard Negation is Prefix, Neg 08 Standard Negation is Tone plus Other Modification, Neg 10 Standard Negation is Infix, Neg 12 Distinct Negation of identity, Neg 13 Distinct Negation of Existence, Neg 14 Distinct Negation of Location, Order N3 01 Demonstrative Adjective Noun, Neg 04 Standard Negation is Suffix, 12 Noun Phrase Adposition} 
from this set of 22 that gives the tree from \namecite{rexova2003cladistic}, see Figure~\ref{FigRexovaTree}.
 
Adding any other parameter to this set of $17$, or removing too many, makes the tree approach the Bouckaert et al tree. This shift suggests that there is an influential (in the sense of being one of the few discriminants of pairs in the family) set of parameters that does not behave like the rest. This indicates that the requirement of identical evolution does not hold. Such anomalies are precisely of interest to the linguists studying syntax.
 
\begin{figure}
    $T_2=$ \Tree [ Hittite [ Tocharian [ Albanian [ Armenian  Greek ]  ] ] ]
    \caption{\namecite{RWT} tree based on morphological data} 
\end{figure}
 
The \namecite{rexova2003cladistic} construction is based on lexicographic data, like the tree of \namecite{bouckaert}, while the
tree of \namecite{RWT} includes both lexicographic and morphological data. None of these previous analyses, with the exception of \namecite{SOBM}, are based on syntactic parameters, so the syntactic input can be seen as an independent verification. We find that the tree of Rexova et al.~lies between the tree of Bouckaert et al., and that of Ringe et al.~when evaluated with phylogenetic invariants: \namecite{SOBM} compares $T_1$ of \namecite{bouckaert} and $T_2$ \namecite{RWT}, computing $T_3$ and $T_4$ invariants (on rolled back parameters) yields 
 
 
$$ \| \Phi_{T_1}(P) \|_{\ell_\infty} = \frac{8}{1331}\approx 0.0060 , \| \Phi_{T_1}(P) \|_{\ell_1} = \frac{61}{2662}\approx 0.0229  . $$
$$ \| \Phi_{T_2}(P) \|_{\ell_\infty} = \frac{8}{1331} \approx 0.0060 , \| \Phi_{T_2}(P) \|_{\ell_1} = \frac{18}{1331} \approx 0.0135. $$
$$ \| \Phi_{T_3}(P) \|_{\ell_\infty}  \approx 0.0060 , \| \Phi_{T_3}(P) \|_{\ell_1} \approx 0.0185. $$
$$ \| \Phi_{T_4}(P) \|_{\ell_\infty}  \approx 0.0060 , \| \Phi_{T_3}(P) \|_{\ell_1} \approx 0.0199. $$

Thus, phylogenetic invariants still favor the tree of \namecite{RWT}. There is an interesting point here,  
in the discrepancy between the result of phylogenetic invariants computation, which is directly based on a geometric formulation of the Markov model hypothesis, and the Markov model based logdet tree that diverges from \namecite{RWT}. 
As very few syntactic parameters separate Hittite and Tocharian, statistical inference for any model, not just the Markov model, becomes difficult. This difficulty in placing closely related languages is a theme that we explore further in the romance family in the following section.  
 
However, note that if the trees appearing were completely random, then one would class that as noise and inherent instability due to sparseness of data, but with support in literature it appears to be more interesting, and suggests that same signals that appear in other data are present here as well; particularly that syntactic structures data are consistent with combined lexicographical and morphological data. 
Next, we try to explore the presence of highly influential structures that seem to be hinted at as noted previously.
 
\subsubsection{Influence analysis}\label{influence_analysis_early_ie}
 
\begin{figure}[H]
{\scriptsize
\includegraphics[scale=0.5]{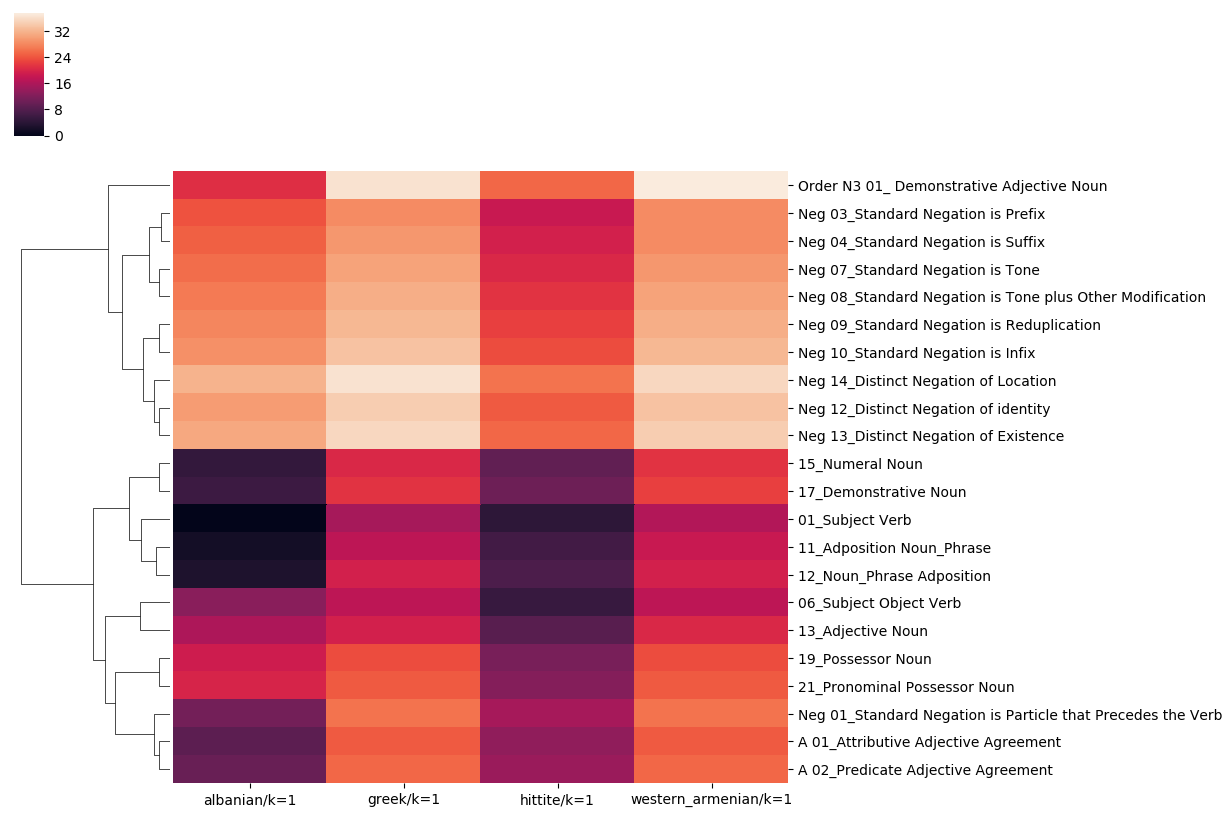}
}
\caption{The clustered heatmap showing how parameters group based on their influences: a cluster of parameters with 
high influence separates out from the rest, for $k = 3$. \label{heatfig} \label{td}}
\end{figure}

To explore the effect of linguistic parameters on determining how distinct each language is from the others in the group we do an influence analysis. A tree is thought of as a boolean function by fixing a root and then considering each of the subsets of the leaves that appear under each interior vertex. We are interested in quantifying how likely the tree is to change on applying \emph{noise} to a parameter: we pick a random set $S$ of parameters, including the parameter of interest, $i$; given the parameter vector $v_l=(v_{l,x})_x$ for a language $l$, 
we flip all coordinates of $v_l$ that are in $S$ to obtain a new vector $v_l^S$. We bound the size of the noise sets to have intersection of size at most $k$ with the sets $\{x:v_{l,x}=1\}$ and $\{x:v_{l,x}=0\}$, so as not to wash out the parameter of interest $i$. We define the influence of $i$ by \begin{equation*} \eta^k_i = \sum_{S\in \Gamma^k_i:\textsc{Tree}[v] \neq \textsc{Tree}[v_S]} \dfrac{1}{|S|}\end{equation*} where $\Gamma^k_i=\{S\,:\,i\in S,\, \max(|S\cap \{x:v_{l,x}=0\}|, |S\cap \{x:v_{l,x}=1\})\leq k \}$. The normalization by $|S|$ adjusts the contribution towards the sensitivity of $i$. We tabulate the parameters which carry largest sensitivity for each member $l$ of the family and for varying $k$ (Figure~\ref{heatfig}); the influences (where we think of highly sensitive parameters as having a higher influence) can also be used to collect parameters to which the family is more sensitive, see~\ref{td}. 
 
Because Hittite and Tocharian are separated by only one parameter, we only consider Hittite in this analysis; the large similarity between the two will not give any meaningful insight. 
 
\begin{figure}[H]
{\scriptsize

\begin{tabular}{lrrrr}
\toprule
{} &  Albanian &  Greek &  Hittite &  wt\_Armenian \\
\midrule
01\_Subject Verb                                    &       0.0 &   15.5 &      4.5 &         16.5 \\
06\_Subject Object Verb                             &      13.0 &   17.5 &      5.5 &         17.5 \\
11\_Adposition Noun\_Phrase                          &       2.0 &   17.5 &      6.5 &         18.5 \\
12\_Noun\_Phrase Adposition                          &       3.0 &   19.5 &      7.5 &         19.5 \\
13\_Adjective Noun                                  &      16.0 &   19.5 &      8.5 &         20.5 \\
15\_Numeral Noun                                    &       5.0 &   20.5 &      9.5 &         21.5 \\
17\_Demonstrative Noun                              &       6.0 &   21.5 &     10.5 &         22.5 \\
19\_Possessor Noun                                  &      19.0 &   23.5 &     11.5 &         23.5 \\
21\_Pronominal Possessor Noun                       &      20.0 &   24.5 &     12.5 &         24.5 \\
A 01\_Attributive Adjective Agreement               &       9.0 &   24.5 &     13.5 &         24.5 \\
A 02\_Predicate Adjective Agreement                 &      10.0 &   25.5 &     14.5 &         25.5 \\
Neg 01\_Standard Negation is Particle that Prece... &      11.0 &   26.5 &     15.5 &         26.5 \\
Neg 03\_Standard Negation is Prefix                 &      24.0 &   28.5 &     18.5 &         28.5 \\
Neg 04\_Standard Negation is Suffix                 &      25.0 &   29.5 &     19.5 &         28.5 \\
Neg 07\_Standard Negation is Tone                   &      26.0 &   30.5 &     20.5 &         29.5 \\
Neg 08\_Standard Negation is Tone plus Other Mod... &      27.0 &   31.5 &     21.5 &         30.5 \\
Neg 09\_Standard Negation is Reduplication          &      28.0 &   32.5 &     22.5 &         31.5 \\
Neg 10\_Standard Negation is Infix                  &      29.0 &   33.5 &     23.5 &         32.5 \\
Neg 12\_Distinct Negation of identity               &      30.0 &   34.5 &     24.5 &         33.5 \\
Neg 13\_Distinct Negation of Existence              &      31.0 &   35.5 &     25.5 &         34.5 \\
Neg 14\_Distinct Negation of Location               &      32.0 &   36.5 &     26.5 &         35.5 \\
Order N3 01\_ Demonstrative Adjective Noun          &      21.0 &   36.5 &     25.5 &         37.5 \\
\bottomrule
\end{tabular}
 
}
\caption{The table of influences with $k=1$. \label{ta}}
\end{figure}

\begin{figure}[H]
{\scriptsize

\begin{tabular}{lrrrr}
\toprule
{} &  Albanian &       Greek &     Hittite &  wt\_Armenian \\
\midrule
01\_Subject Verb                                    &      26.0 &   86.166667 &   52.500000 &        209.0 \\
06\_Subject Object Verb                             &     127.0 &  107.500000 &   53.500000 &        210.0 \\
11\_Adposition Noun\_Phrase                          &      28.0 &   88.166667 &   54.500000 &        211.0 \\
12\_Noun\_Phrase Adposition                          &      29.0 &  109.500000 &   55.500000 &        212.0 \\
13\_Adjective Noun                                  &     130.0 &   90.166667 &   56.500000 &        213.0 \\
15\_Numeral Noun                                    &      31.0 &   91.166667 &   57.500000 &        214.0 \\
17\_Demonstrative Noun                              &      32.0 &   92.166667 &   58.500000 &        215.0 \\
19\_Possessor Noun                                  &     133.0 &  113.500000 &   59.500000 &        216.0 \\
21\_Pronominal Possessor Noun                       &     134.0 &  114.500000 &   60.500000 &        217.0 \\
A 01\_Attributive Adjective Agreement               &      35.0 &   95.166667 &   61.500000 &        217.0 \\
A 02\_Predicate Adjective Agreement                 &      36.0 &   96.166667 &   62.500000 &        218.0 \\
Neg 01\_Standard Negation is Particle that Prece... &      37.0 &   97.166667 &   63.500000 &        219.0 \\
Neg 03\_Standard Negation is Prefix                 &     138.0 &  118.500000 &  107.166667 &        221.0 \\
Neg 04\_Standard Negation is Suffix                 &     139.0 &  119.500000 &  108.166667 &        221.0 \\
Neg 07\_Standard Negation is Tone                   &     140.0 &  120.500000 &  109.166667 &        222.0 \\
Neg 08\_Standard Negation is Tone plus Other Mod... &     141.0 &  121.500000 &  110.166667 &        223.0 \\
Neg 09\_Standard Negation is Reduplication          &     142.0 &  122.500000 &  111.166667 &        224.0 \\
Neg 10\_Standard Negation is Infix                  &     143.0 &  123.500000 &  112.166667 &        225.0 \\
Neg 12\_Distinct Negation of identity               &     144.0 &  124.500000 &  113.166667 &        226.0 \\
Neg 13\_Distinct Negation of Existence              &     145.0 &  125.500000 &  114.166667 &        227.0 \\
Neg 14\_Distinct Negation of Location               &     146.0 &  126.500000 &  115.166667 &        228.0 \\
Order N3 01\_ Demonstrative Adjective Noun          &      47.0 &  107.166667 &   73.500000 &        230.0 \\
\bottomrule
\end{tabular}
 
}
\caption{The table of influences with $k\leq 2$. \label{tb}}
\end{figure}

\begin{figure}[H]
{\scriptsize

\begin{tabular}{lrrrr}
\toprule
{} &     Albanian &        Greek &     Hittite &  wt\_Armenian \\
\midrule
01\_Subject Verb                                    &  2255.066667 &  2386.733333 &   52.500000 &       1199.0 \\
06\_Subject Object Verb                             &  1793.800000 &  1922.800000 &   53.500000 &       1200.0 \\
11\_Adposition Noun\_Phrase                          &  2257.066667 &  2388.733333 &   54.500000 &       1201.0 \\
12\_Noun\_Phrase Adposition                          &  2258.066667 &  1924.800000 &   55.500000 &       1202.0 \\
13\_Adjective Noun                                  &  1796.800000 &  2390.733333 &   56.500000 &       1203.0 \\
15\_Numeral Noun                                    &  2260.066667 &  2391.733333 &   57.500000 &       1204.0 \\
17\_Demonstrative Noun                              &  2261.066667 &  2392.733333 &   58.500000 &       1205.0 \\
19\_Possessor Noun                                  &  1799.800000 &  1928.800000 &   59.500000 &       1206.0 \\
21\_Pronominal Possessor Noun                       &  1800.800000 &  1929.800000 &   60.500000 &       1207.0 \\
A 01\_Attributive Adjective Agreement               &  2264.066667 &  2395.733333 &   61.500000 &       1252.0 \\
A 02\_Predicate Adjective Agreement                 &  2265.066667 &  2396.733333 &   62.500000 &       1253.0 \\
Neg 01\_Standard Negation is Particle that Prece... &  2266.066667 &  2397.733333 &   63.500000 &       1254.0 \\
Neg 03\_Standard Negation is Prefix                 &  1804.800000 &  1933.800000 &  135.166667 &       1211.0 \\
Neg 04\_Standard Negation is Suffix                 &  1805.800000 &  1934.800000 &  136.166667 &       1256.0 \\
Neg 07\_Standard Negation is Tone                   &  1806.800000 &  1935.800000 &  137.166667 &       1257.0 \\
Neg 08\_Standard Negation is Tone plus Other Mod... &  1807.800000 &  1936.800000 &  138.166667 &       1258.0 \\
Neg 09\_Standard Negation is Reduplication          &  1808.800000 &  1937.800000 &  139.166667 &       1259.0 \\
Neg 10\_Standard Negation is Infix                  &  1809.800000 &  1938.800000 &  140.166667 &       1260.0 \\
Neg 12\_Distinct Negation of identity               &  1810.800000 &  1939.800000 &  141.166667 &       1261.0 \\
Neg 13\_Distinct Negation of Existence              &  1811.800000 &  1940.800000 &  142.166667 &       1262.0 \\
Neg 14\_Distinct Negation of Location               &  1812.800000 &  1941.800000 &  143.166667 &       1263.0 \\
Order N3 01\_ Demonstrative Adjective Noun          &  2276.066667 &  2407.733333 &   73.500000 &       1220.0 \\
\bottomrule
\end{tabular}
 
}
\caption{The table of influences with $k\leq 3$. \label{tc}}
\end{figure}
 
Based on the influences that the syntactic structures carry, it is apparent that there are two classes of 
syntactic structures that are distinct in how much information they encode about the structure of the family. 
 
In particular, note how the top half of the tree in Figure~\ref{heatfig} consists entirely of negation features,
which are set to zero on languages where negation is a separate word and not expressed via prefix, suffix, 
or infix. So it is clearly not surprising that those features are not significant in comparison
to the one negation feature (negation as a particle) that is present in those languages. 
 
 
 
\subsection{Romance languages}\label{sswl_romance}

For the romance languages we obtain the following tree, which we have drawn with
the root placed next to Latin: 
 
\begin{figure}[H]
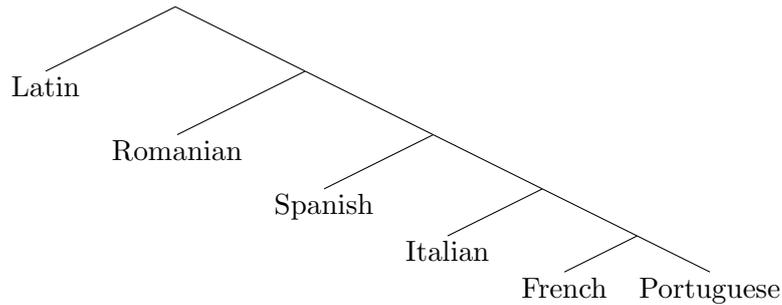

\begin{center}
\Tree[  Latin [  Romanian [ Spanish [ Italian [ French Portuguese ] ] ] ] ]  
\end{center}
\caption{The \logdetnj Romance tree using 85 parameters from across SSWL and Longobardi completely set for the family.\label{rom_logdet_tree}}
\end{figure}


This tree misplaces Spanish. On moving Spanish in proximity of Portuguese, 
we obtain the historically accepted tree:
 
\begin{figure}[H]
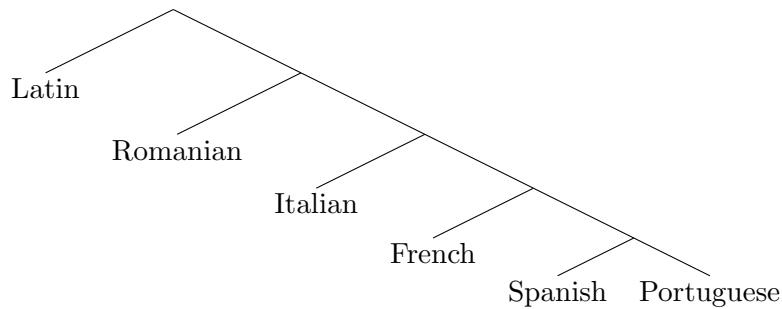

\begin{center}
\Tree[  Latin  [ Romanian [ Italian  [ French [ Spanish Portuguese ] ] ] ] ]  
\caption{The historically accepted tree for the Romance family}
\end{center}
\end{figure}
 
The logdet matrix correctly shows that Spanish is indeed closest to Portuguese. However, the raw logdet value is not what neighbor joining optimizes, so the expected (Spanish, Portuguese) subtree does not emerge. Additionally, the logdet values involving Spanish and Portuguese show high degree of asymmetry. 
 
The similar misplacement of Spanish next to (Romanian, Latin) subtree also appears in \namecite{SAHWM}. We note that geographic proximity, the related history of Portuguese and Spanish, and likely coevolution, makes it a candidate for violating the evolution on tree assumption: we expect these two to have much more similarity and interaction than possible on a tree. \newpara

\begin{figure}
    \begin{center}
    \begin{tabular}{lrrrrrr}
        \toprule
        {} &     Latin &  Romanian &   Italian &    French &   Spanish &  Portuguese \\
        \midrule
        Latin      &  0.000000 &  0.418368 &  0.526778 &  0.686986 &  0.513258 &    0.561157 \\
        Romanian   &  0.418368 &  0.000000 &  0.180652 &  0.237485 &  0.123650 &    0.152100 \\
        Italian    &  0.526778 &  0.180652 &  0.000000 &  0.098833 &  0.047602 &    0.023741 \\
        French     &  0.686986 &  0.237485 &  0.098833 &  0.000000 &  0.100617 &    0.073974 \\
        Spanish    &  0.513258 &  0.123650 &  0.047602 &  0.100617 &  0.000000 &    0.023861 \\
        Portuguese &  0.561157 &  0.152100 &  0.023741 &  0.073974 &  0.023861 &    0.000000 \\
        \bottomrule
        \end{tabular}\caption{Raw logdet values for the Romance family}
\end{center}
\end{figure}

The \logdetnj tree obtained is also not stable when attempting reconstruction after subsampling down to approximately 60\% of the data: it's recovered with probability approximately $1/2$ and the correct topology appears with probability approximately $1/50$ where the probabilities are computed across 1000 trials. 
Computing phylogenetic invariants, we find that the $\Phi^{\ell_\infty}\approx 0.00156$ invariant does not separate the two topologies, but $\Phi^{\ell_1}$ does in fact separate them and surprisingly selects the \logdetnj tree, taking a value $\approx 0.0094$ as versus $\approx 0.0111$ for the historically accepted tree. 
 
\begin{figure}
\includegraphics[scale=0.8]{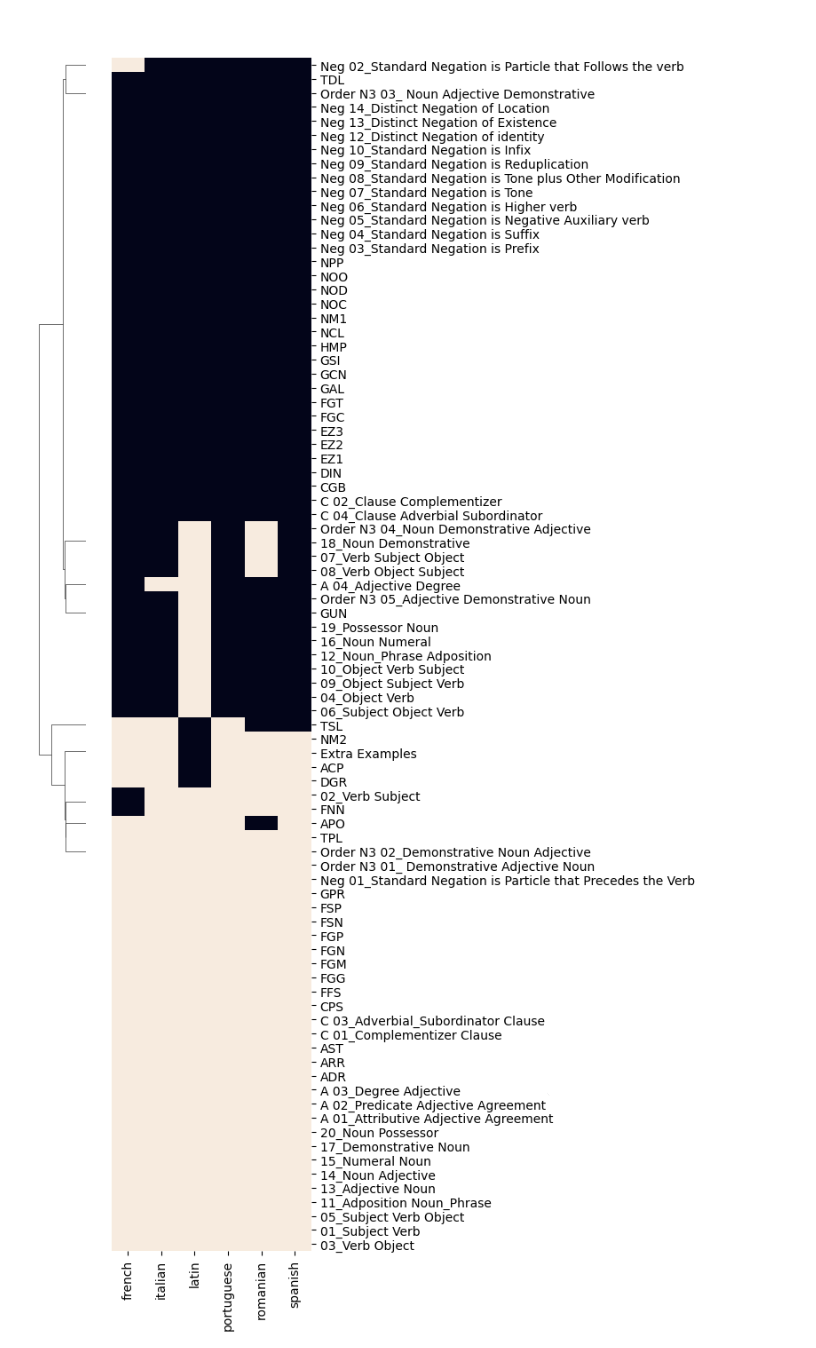}
\caption{Clustering analysis of syntactic features for the Romance languages. (Blue indicates unset parameters.) This can be compared with the cluster analysis of syntactic features over the entire database of languages, as discussed in \namecite{marcolli_topling} and \namecite{marcolli_hk}. The two colors denote the two states of the parameters. \label{fig_romance_param_clusters}}
\end{figure}

A natural question to ask is what parameters are the primary drivers of the ambiguity in the tree reconstruction. Compared to the Early Indo-European languages case, there are more syntactic parameters for which data are available, but at the same time there are very few parameters that separate members of this family. This sparsity, coupled with a larger parameter set, makes the influence analysis techniques used in section \ref{influence_analysis_early_ie} computationally infeasible. However, the simple clustering-based approach of Figure~\ref{fig_romance_param_clusters} is revealing.
 
\subsubsection{Maximum likelihood model}
 
Building the tree using the greedy scheme optimizing covariances, we obtain the following tree that misplaces 
French, Spanish and Italian.
 
\begin{figure}[H]
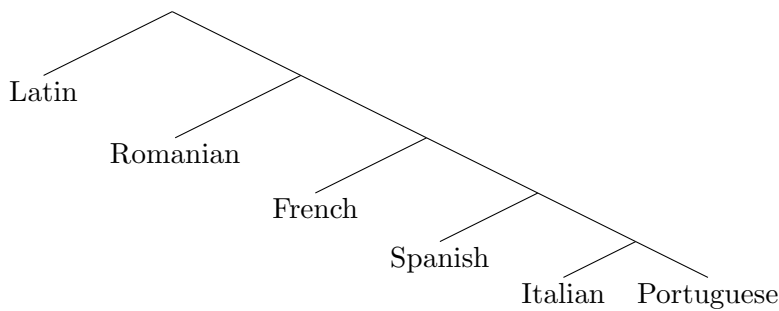

\begin{center}
\Tree[  Latin  [ Romanian [ French  [ Spanish [  Italian Portuguese ] ] ] ] ] 
\caption{Tree based on covariance}
\end{center}
\end{figure}
 
With a comparatively larger set of available parameters, this family lends itself to a richer analysis along the lines we have sketched; we examine this family in more detail. We begin by constructing the maximum likelihood estimate for the tree parameter using the topology of the logdet tree, which would give the correct topology under the assumption that the data come from a general Markov model.\newpara
 
Starting with the matrix of data with one column for each of the $m=85$ syntactic parameters, and each column containing the values the parameter takes at the leaves of the phylogenetic tree for the Romance family (L:Latin, R:Romanian, S:Spanish, I:Italian, F:French, P:Portuguese). Under the assumption that all parameters are independent, the likelihood of the data matrix, $Z=[z^1,z^2\dots z^m]$, is the product of individual likelihoods of each column $z^i$, $$ L(Z) = \prod_{i\in [m]} L(z^i) \, . $$
 
\begin{center}
\begin{figure}[!htb]
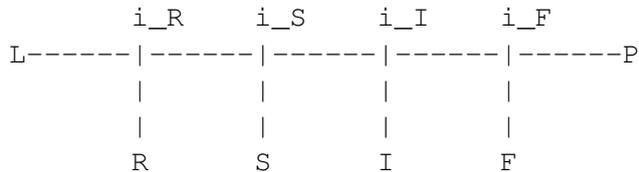

{
\fontfamily{courier}\selectfont
\begin{verbatim}
       i_R    i_S    i_I    i_F
L------|------|------|------|------P
       |      |      |      |
       |      |      |      |
       R      S      I      F
\end{verbatim}
}
\caption{The $\logdetnj$tree of Figure~\ref{rom_logdet_tree} with $\{L, R, S, I, F, P, i_R, i_S, i_I, i_L\}$ labeling. \label{romance}}
\end{figure}
\end{center}

The likelihood for a column of data for the \logdetnj tree model can be computed by assigning
a probability distribution over $\{0, 1\}$ parameter values at the root, $\pi=[\pi_0, 1-\pi_0]^T$, and 
2-dimensional transition matrices, $$M_k=\begin{bmatrix} \theta^0_k & 1-\theta^0_k \\  1-\theta^1_k & \theta^1_k \end{bmatrix}$$ to each edge (the pendant edges are labelled by leaves, the interior edges by the left vertex) for $k\in \{L, R, S, I, F, P, i_R, i_S, i_I, i_L\}$ in the L 
rooted tree of Figure~\ref{romance}.
 
For a given vector $z\in \{0, 1\}^5$, we are interested in the probability of seeing $z$ at the leaves as a function of the parameters: $P(z)\equiv P(R=z_1, S=z_2\dots P=z_5) =f(M_k, \pi)$. The likelihoods can be computed using dynamic programming but here we explicitly sum over the internal states at $i_l, l \in \{R, S, I, F\}\equiv \text{Int}$ and the state $x$ at the root. For simplicity we use the identification $R=1, S=2, I=3, F=4, P=5$ i.e. $M_R=M_1$ etc. 
 
$$P(z) = \sum_{x\in\{0,1\}}\sum_{v\in \{0,1\}^4}P(z|i_k = v_k, k\in [4],x)P(v,x) $$
 
The probabilities at the leaves can be read off the transition matrices once the interior states are fixed (where we use independence). This gives:
$$P(z|i_k = v_k,k\in [4],x) = M_R[v_1,z_1]\cdot M_S[v_2,z_2]\cdot M_I[v_3,z_3]\cdot M_F[v_4,z_4]\cdot M_P[v_4, z_4]$$
 
All that remains to estimate is $P(v,x)$. This follows similarly, since $$ P(v,x) = P(v_4|v_3)P(v_3|v_2)P(v_2|v_1)P(v_1,x) = M_{i_I}[v_3, v_4]\cdot M_{i_S}[v_2, v_3]\cdot M_{i_R}[v_1, v_2]\cdot P(v_1, x),$$
where we used that $v_i$ only depends on $v_{i-1}$. Now $P(v_1, x) = P(v_1|x)P(x)= M_L[x, v_1]\pi[x]$, so that 
$$P(z)=\sum_{x, v} \prod_{i\in[4]} M_i[v_i,z_i]\cdot M_P[v_4, z_4]\cdot M_{i_I}[v_3, v_4]\cdot M_{i_S}[v_2, v_3]\cdot M_{i_R}[v_1, v_2]\cdot M_L[x, v_1]\cdot \pi[x]\,  .$$  
 
This will give likelihood of the data as a polynomial in the model parameters after taking the product over all parameters. We maximize this using gradient descent to get the maximum likelihood estimate for the parameters and simulated 10000 evolutions on this tree. Collecting into groups of 85 (which is the number of parameters used for the \logdetnj tree), this gives approximately 110 trials. Since the simulated evolutions are independent, the distribution of distances between any two leaves for all trials are identical; their distribution approaches a Gaussian by the central limit theorem.\footnote{This convergence can be quantified with the Berry--Esseem theorem, see \namecite{durrett2019probability}.} For the leaves $i,j$, we used the estimated standard deviation, $\sigma^{ij}$,  and mean, $\mu^{ij}$, from the simulated data to assign a z-score ($z^{ij}=(d^{ij} - \mu^{ij})/\sigma^{ij}$) to the distances seen in the actual data, $d^{ij}$. The distance between pairs are not independent, so coalescing the data into single statistics is not straightforward, and as the table of z-scores demonstrates, different subtrees within the tree behave differently; for example, any distances computed between the subgroup of (French, Portuguese) and the rest of the family are overestimated by more than two standard deviations in the model, while other distances are within tolerance (one standard deviation of the mean). This gives support to the idea of there being more interaction in this family than tree models permit. 
 
\begin{table}
\begin{center}
\begin{tabular}{lllllll}
    \toprule
    {} & l1 & l2 &       mean &        std &          z &          d \\
    \midrule
    I/P &  I &  P &    2.78336 &   0.893444 &   -3.08875 &  0.0237408 \\
    F/I &  F &  I &    2.87366 &   0.946365 &   -2.93209 &  0.0988331 \\
    F/S &  F &  S &    2.85387 &   0.973022 &   -2.82959 &   0.100617 \\
    F/R &  F &  R &     2.8915 &   0.997764 &   -2.65996 &   0.237485 \\
    P/R &  P &  R &    2.90869 &    1.04035 &   -2.64968 &     0.1521 \\
    P/S &  P &  S &    2.92124 &    1.12043 &   -2.58596 &  0.0238608 \\
    F/L &  F &  L &    2.81107 &    1.01573 &   -2.09119 &   0.686986 \\
    L/P &  L &  P &    2.84234 &    1.22816 &    -1.8574 &   0.561157 \\
    R/S &  R &  S &   0.131869 &  0.0597361 &  -0.137579 &    0.12365 \\
    F/P &  F &  P &  0.0765047 &  0.0436683 & -0.0579418 &  0.0739745 \\
    I/L &  I &  L &   0.520308 &   0.129903 &  0.0498074 &   0.526778 \\
    I/R &  I &  R &   0.172632 &  0.0677208 &    0.11843 &   0.180652 \\
    L/R &  L &  R &    0.40522 &   0.104813 &   0.125442 &   0.418368 \\
    I/S &  I &  S &  0.0420906 &   0.033077 &   0.166614 &  0.0476017 \\
    L/S &  L &  S &   0.479633 &   0.122479 &   0.274532 &   0.513258 \\
    \bottomrule
    \end{tabular}
\end{center}\caption{z-score table for actual logdet distances compared to the maximum likelihood model}
\end{table}
 
\begin{figure}
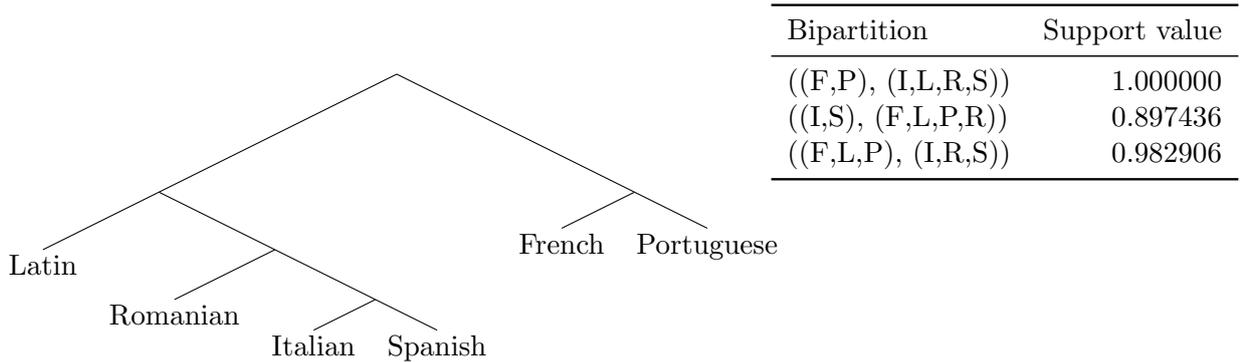

\begin{subfigure}[b]{0.6\textwidth}
\begin{center}
\Tree[  [ Latin   [ Romanian  [ Italian  Spanish ]  ]  ]    [ French  Portuguese ]  ]
\end{center}
\end{subfigure}
\begin{subfigure}[b]{0.3\textwidth}
\begin{tabular}{lr}
\toprule
Bipartition &  Support value \\
\midrule
((F,P), (I,L,R,S))  &      1.000000 \\
((I,S), (F,L,P,R))  &      0.897436 \\
((F,L,P), (I,R,S))  &      0.982906 \\
\bottomrule
\end{tabular}
\end{subfigure}
\caption{Covariance based tree using simulated data from the maximum likelihood model. A bipartition support value corresponding to the probability that a bipartition is observed in simulated data is tabulated for the non-trivial bipartitions. (The root placement here is only for graphical convenience, while the historical root should be placed next to Latin.) \label{FigTreeCovar}}
\end{figure}

Using the simulated data across 1000 simulations, we again build the tree based on covariance, see Figure~\ref{FigTreeCovar}.  
 
Again, this tree reports incorrectly the relative positions of French, Italian, and Spanish, with
French and Spanish interchanged with respect to the tree considered historically correct.
The fact that the greedy scheme places French closer to Portuguese is a significant misplacement 
because it represents that Portuguese and French have higher covariance than Spanish under the assumption on the Markov model, suggesting that this does not describe the syntactic evolution well. 
A pattern showing that the newer Romance languages are often conflated in varied ways 
has emerged: the misplacements primarily involve French, Portuguese, and Spanish.
 
These languages share a history that suggests that the limited interaction possible on a tree fails to describe them. The close linguistic relatedness also makes it likely that the syntactic structures are not completely independent. This should be compared, for instance, with the analysis in Sections~4.4 and 4.5 of \namecite{marcolli_topling}, where the dimensionality of the space of syntactic parameters is analyzed over different linguistic families, showing a drop in dimension in certain families, that corresponds to the presence of additional family-specific relations (see Figure~13 of \namecite{marcolli_topling} for the case of the Romance languages).
 
We move to consider both the sufficiency of the tree topology and independence and identical evolution assumption on the syntactic structures. 

 \subsubsection{Independent and identical evolution assumption}
 
Because of their geographic proximity, French, Spanish and Portuguese are likely to have experienced
a higher degree of interaction, including at the syntactic level, than what normally expected in a Markov
model on a tree (see our general discussion at the beginning of the paper on the model and its
assumptions). It also appears that the syntactic structures within the Romance languages, 
especially when compared to Latin, are more highly correlated than what expected in general in
terms of the dimensionality of the space of syntactic parameters (see Sections~4.4 and 4.5 of \namecite{marcolli_topling}).
The more recent evolution of the modern Romance languages also contributes to make their syntactic parameters  less likely to behave like independent/identically evolving. Since we do have the simulated data from the maximum likelihood model, we can compare the simulated against the actual, and indeed, we find that this is exactly what we observe. 
 
The possibility of convergent evolution rather than interaction leading to the observed syntactic structure of the Romance languages is discussed in \namecite{Lo12}. Other possibilities, beyond interaction and convergent evolution, include the possibility of long branch attraction due to the languages splitting off from the trunk starting at Latin in a sort of sequential fashion.
A more refined model, such as a more detailed formulation of the infinite sites model, will be needed to distinguish these hypotheses. For the purpose of the present paper, we provide the following analysis.
We start by randomly ordering the syntactic parameters in a given set $S$. Then for each triple of languages, $L=\{ l_1, l_2, l_3 \}$, we consider the probability that the syntactic parameter, $s_i$ is set in all three, as well as the probability that the next one in the given ordering, $s_{i+1}$, is also set. We compute:
$$\dfrac{\Pr_S[\forall l\in L, s_i=1]^2}{1/|S|+ \Pr_S[\forall l\in L, s_i=1 \wedge s_{i+1}=1]}\, . $$
 
The choice $|L|=3$ is made because this is the largest size for which the size of the state space, $2^3$ is ``small'' compared the the number of syntactic structures that are available. 
 We follow this by computing the same for the simulated data. Averaging over 100 random orderings of $S$, we consider the $z$-score for each triple of languages against the distribution from the simulated data. Since the simulated data come from iid evolutions on the maximum likelihood model, they give the expected distribution for what the statistics for iid evolution looks like. This is summarized by the following Table \ref{tablezscores}.

\begin{table}
\begin{center}
\begin{tabular}{lr}
\toprule
{} &   $z$-score \\
\midrule
F:S:P &  0.222647 \\
I:F:P &  0.192141 \\
I:F:S &  0.199228 \\
I:S:P &  0.180108 \\
L:F:P &  0.102739 \\
L:F:S &  0.245780 \\
L:I:F &  0.223499 \\
L:I:P &  0.213800 \\
L:I:S &  0.036244 \\
L:R:F &  0.167095 \\
L:R:I & -0.004172 \\
L:R:P &  0.149725 \\
L:R:S & -0.004571 \\
L:S:P &  0.233173 \\
R:F:P &  0.159186 \\
R:F:S &  0.210917 \\
R:I:F &  0.206196 \\
R:I:P &  0.183321 \\
R:I:S &  0.041099 \\
R:S:P &  0.188342 \\
\bottomrule
\end{tabular}
\end{center}\caption{The language triples and associated $z$-score for Latin (L), Spanish (S), French (F), Portuguese (P) and Romanian (R)) \label{tablezscores}}
\end{table}
 
In general, the statistics match what is expected from the simulated data, as it all lies within a quarter standard deviation. This implies that the statistics we will compute next, to test the adequacy of the tree topology, are largely unconfounded by the deviation from iid evolution. 
 
\subsubsection{Subfamily splits against maximum likelihood model}
 
The placement of Portuguese and Spanish is one of the confounding factors that repeatedly appears. We consider the partitions of the family where Portuguese and Spanish are separated, as well as where they form a cherry (while keeping together the consistent Romanian/Latin pair). If these partitions reflect genuine splits in the data, then the rank of the flattening for these partitions must be 2. The distance, $d$, to the closest rank 2 matrix to the flattening matrix can be computed as the norm of its singular values vector after excluding the top 2. We compute this for the simulated maximum likelihood data, $d_{sim}$ where the logdet tree topology implies that Portuguese and Spanish separate. The $z$-score for the actual data $d$ value, $d_{actual}$, against the background from the simulated data is tabulated as in Table \ref{tablezscores2}.
 
\begin{table} 
\begin{center}
\begin{tabular}{lrrrr}
\toprule
{} &  d\_actual &  mean[d\_sim] &  std\_dev[d\_sim] &   z-score \\
\midrule
F,P,S; L,R,I &  0.012677 &     0.035968 &        0.013168 & -1.768730 \\
I,F,P,S; L,R &  0.012585 &     0.031150 &        0.013502 & -1.375018 \\
I,F,P; L,S,R &  0.012274 &     0.034641 &        0.013497 & -1.657230 \\
I,F,S; L,P,R &  0.017000 &     0.220083 &        0.022537 & -9.010975 \\
P,S; F,L,R,I &  0.011765 &     0.223664 &        0.023124 & -9.163532 \\
\bottomrule
\end{tabular}
\end{center}\caption{Table of $d$ values and $z$-scores for splittings of the tree. \label{tablezscores2}}
\end{table}
 
The table shows that the $d$ value implied by the split from the logdet tree, [[L, R, S], [I, F, P]] lies one variance outside what is expected from the maximum likelihood model. And the [S, P] cherry comes with an order of magnitude more extreme $z$-score. This corresponds to data reflecting that Spanish and Portuguese are extremely likely to form a cherry than what is expected by the iid evolution on the tree, because the distance to the nearest rank 2 approximation is much smaller than what is obtained from the maximum likelihood model. Overall, the data is at least one variance outside what is expected. The Portuguese--Spanish cherry statistics strongly suggests that the tree topology is not capturing the full range of interactions.

\section{The logdet phylogenetic signal}

The richness of the data of \namecite{longo2020} offers a way to test the scales at which the evolutionary models continue to be reliable beyond the setting of small-scale examples considered previously. We compare the phylogenetic tree reconstructed using logdet/general Markov and modified Jaccard index/infinite site model and discuss them in context of the results of \namecite{longo2020} using Bayesian phylogenetics (as implemented in BEAST software package, \namecite{beast2014}) as well as UPGMA clustering with modified Jaccard index. 
 
Unlike the LanGeLin and SSWL dataset where we only use the parameters that are independent and known in all languages for which the construction is being carried out, for the \namecite{longo2020} dataset, we use all parameters, only restricting to independent parameters when computing pairwise distances. This means the distance between different pairs may be based on a different set of parameters. This is done following \namecite{longo2020}, as otherwise we do not have enough parameters if we discard all that are not independent for any language. Under the assumption that all sites are i.i.d., this does not make a difference. For the reconstructed trees, because the linguistic evolutionary processes are not necessarily memoryless, the branch lengths are not meaningful; and we will reroot the trees with input from what is commonly agreed upon in linguistics literature. To conclude this section, we will make a quantitative note of the difference between modified Jaccard and logdet based trees using Robinson--Fould metric, but before that we undertake a qualitative analysis. 
 
 \subsection{Languages not included in SSWL and Longobardi datasets}
 
We first consider language families that are not included in the SSWL and Longobardi datasets.
\begin{itemize}
\item The Indo-Iranian family is represented in the Ceolin dataset by Marathi, Hindi, Pashto.
We see here an example of misplacement across language families, as the
non-Indo-European Dravidian languages Telugu and Tamil are placed inside the
Indo-Iranian subtree of the Indo-European tree. 
This appears problematic, since the Dravidian and Indo-Iranian languages do not share a common
Ancestor that can be recovered. However, note that here we are only evaluating distances, not an evolutionary model. 
    
\begin{itemize}
\item Indo-Iranian and Dravidian languages distance relations reconstructed using the Logdet metric: 
$$
\Tree[ Marathi  [ Hindi  [ Telugu Tamil ]  ]  Pashto ]
$$
\item Indo-Iranian and Dravidian languages distance relations reconstructed using the modified Jaccard metric:
$$\Tree[  [ Telugu Tamil ]   [ Marathi Hindi ]  Pashto ]$$
\end{itemize}
 
The modified Jaccard index gives a more accurate reconstruction with the pairs Tamil/Telugu and Hindi/Marathi correctly identified, with the Tamil/Telugu forming a separate grouping of the two Dravidian languages, identical to the results of \namecite{longo2020}.  
 
\item East Asian languages, Korean, Japanese, Cantonese and Mandarin are correctly reconstructed using both and are in agreement with \namecite{longo2020} $$\Tree[ [ Korean Japanese ] [ Cantonese Mandarin ] ]$$
 
\end{itemize}
 
\subsection{Greco-Romance Languages}\label{GrecoRomanceSec}
 
Data for a superset of the Greco-Romance languages considered in section \ref{sswl_longo_analysis} is available in the \namecite{longo2020} dataset. 
We group the set of Hellenic and Romance languages because the data of syntactic features include the microvariations of a family of both Romance and Hellenic dialects of Southern Italy (see the specific analysis of \namecite{microvar}).
 
Restricting to just the Greco-Romance languages in the Ceolin data, both logdet and modified Jaccard similarity give identical trees; this was expected based on the high degree of asymmetry in maximum likelihood model. Portuguese and Spanish are now correctly placed. We do see a misplacement of two Italian dialects: Parma and Casalesco, but we note that this misplacement is also present in the BEAST trees of \namecite{longo2020}, and that it disappears when we consider the full Indo-European family instead of just the Greco-Romance family. This suggests that this is possibly arising from the large sampling bias in this set which contains a disproportionate number of closely related Italian dialects 
\begin{figure}[H]
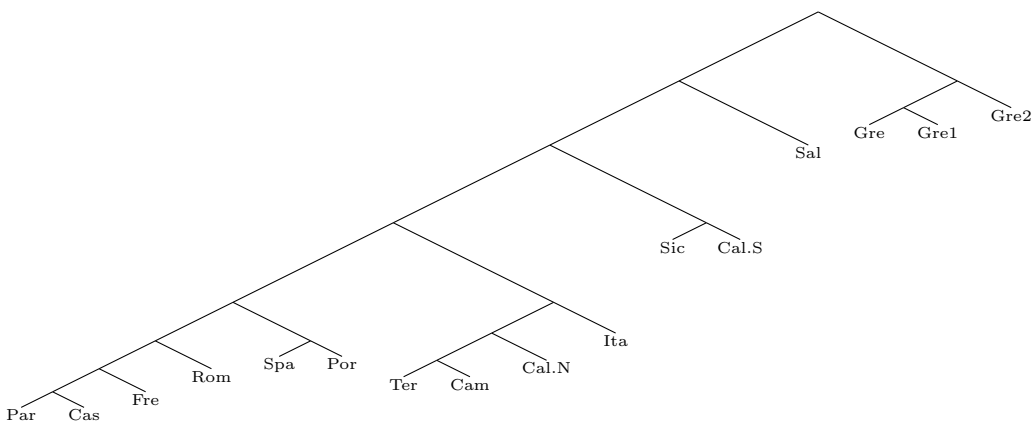

{\tiny
$$\Tree[ [  [  [  [  [  [  [ Par Cas ]  Fre ]  Rom ]   [ Spa Por ]  ]   [  [  [ Ter Cam ]  Cal.N ]  Ita ]  ]   [ Sic Cal.S ]  ]  Sal ]
[  [    Gre   Gre1 ]  Gre2 ] ]$$
}\caption{The Greco-Romance tree obtained from \logdetnj construction that is identical to the modified Jaccard index tree when restricted to the Greco-Romance languages: Romanian (Rom), French (Fre), Spanish (Spa), Portuguese (Por), Italian (Ita), Parma (Par), Casalasco (Cas), 
Teramano (Ter), Campano (Cam), Calabrese-Northern (Cal.N), Calabrese-Southern (Cal.S), Sicilian (Sic), Salentino (Sal), Greek (Gre), 
Greek-Calabria N.1 (Gre1), Greek-Calabria N.1 (Gre2); Note the tree lengths are not to scale; we are considering tree topology only.}
\end{figure}
 
As noted, this tree carries some misplacements. In view of these inaccuracies, we are led to conclude that either sampling biases or failure of assumptions underlying the models are significant. One notes that Greek and Romance families are quite distinct, and one cannot suppose that a hypothetical root from which these evolve can be recovered from the data. We will reconsider the Greco-Romance languages in the context of the full Indo-European family in section \ref{full_ie_family}. 
 
As a final check, we revisit the SSWL-LanGeLin data for the Romance family from section \ref{sswl_romance}, and apply the modified Jaccard index construction to see if the change in the model resolves the persistent issues there. We find that the reconstruction is identical, with modified Jaccard index values similar to the logdet values. 
 
\begin{table}[H]
\begin{center}
\begin{tabular}{lrrrrrr}
\toprule
{} &    French &   Italian &     Latin &  Portuguese &  Romanian &   Spanish \\
\midrule
French     &  0.000000 &  0.098833 &  0.686986 &    0.073974 &  0.237485 &  0.100617 \\
Italian    &  0.098833 &  0.000000 &  0.526778 &    0.023741 &  0.180652 &  0.047602 \\
Latin      &  0.686986 &  0.526778 &  0.000000 &    0.561157 &  0.418368 &  0.513258 \\
Portuguese &  0.073974 &  0.023741 &  0.561157 &    0.000000 &  0.152100 &  0.023861 \\
Romanian   &  0.237485 &  0.180652 &  0.418368 &    0.152100 &  0.000000 &  0.123650 \\
Spanish    &  0.100617 &  0.047602 &  0.513258 &    0.023861 &  0.123650 &  0.000000 \\
\bottomrule
\end{tabular}\caption{Table for logdet metric}
\end{center}
\end{table}
 
\begin{table}[H]
\begin{center}
\begin{tabular}{lrrrrrr}
\toprule
{} &    French &   Italian &     Latin &  Portuguese &  Romanian &   Spanish \\
\midrule
French     &  0.000000 &  0.100000 &  0.415094 &    0.076923 &  0.209302 &  0.102564 \\
Italian    &  0.100000 &  0.000000 &  0.346154 &    0.025641 &  0.162791 &  0.051282 \\
Latin      &  0.415094 &  0.346154 &  0.000000 &    0.365385 &  0.294118 &  0.352941 \\
Portuguese &  0.076923 &  0.025641 &  0.365385 &    0.000000 &  0.142857 &  0.026316 \\
Romanian   &  0.209302 &  0.162791 &  0.294118 &    0.142857 &  0.000000 &  0.121951 \\
Spanish    &  0.102564 &  0.051282 &  0.352941 &    0.026316 &  0.121951 &  0.000000 \\
\bottomrule
\end{tabular}\caption{Table for modified Jaccard index}
\end{center}
\end{table}
 
\subsection{Germanic Languages}
 
For the Germanic family, the \namecite{longo2020} data contain the additional North Germanic languages Danish and Norwegian, and are missing Swedish. Using Icelandic as the outgroup to root the rest we correctly recover the North/West split with both logdet and modified Jaccard. The interior structures differ, with modified Jaccard placing Faroese with Danish, where the expected would be Norwegian, while lodget fails to assign Dutch and Afrikaans together (same as UPGMA from \namecite{longo2020}, which also fails to recover West and North Germanic split, while Ceolin et al's BEAST reconstruction places German next to Afrikaans).   
\begin{figure}[H]
$$\Tree[ [  [  [ German Dutch ]  Afrikaans ]  English ] [  [       [ Norwegian Faroese ]  Danish   ]  Icelandic ] ]$$
\caption{\logdetnj construction}
\end{figure}
\begin{figure}[H]
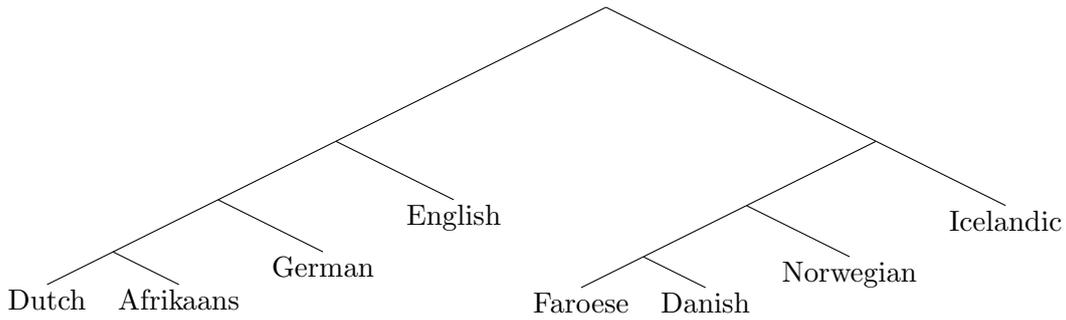

$$\Tree[ [  [  [ Dutch Afrikaans ]  German ]  English ] [       [  [ Faroese Danish ]  Norwegian ]    Icelandic ] ]$$
\caption{Modified Jaccard construction}
\end{figure}
 
 
 
 
 
\subsection{Balto-Finnic, Ugric and Altaic}
 
For the Balto-Finnic, Ugric and Altaic languages the logdet and modified Jaccard both give very similar structures, with some disagreement between the two pairs of closely related dialects of Mari (Mar1, Mar2) and Udmurt (Udm1, Udm2). Similar conflation is observed in the BEAST tree from \namecite{longo2020}. Compared to the LanGeLin \logdetnj construction, we correctly recover the relationship between Estonian (Est) and Finnish (Fin), between Khanty (Kha1, Kha2) and Hungarian (Hun), and that between Turkish (Tur) and Yakut (Yak). We recover the relations between Evenki (Eve), Uzbekh (Uzb), Yakut and Turkish that are the same as Ceolin et al's as well. The modified Jaccard tree provides a better reconstruction as it separates the three Turkic languages -- Uzbek, Yakut and Turkish into their own subtree within the Altaic languages. 
 
\begin{figure}[H]
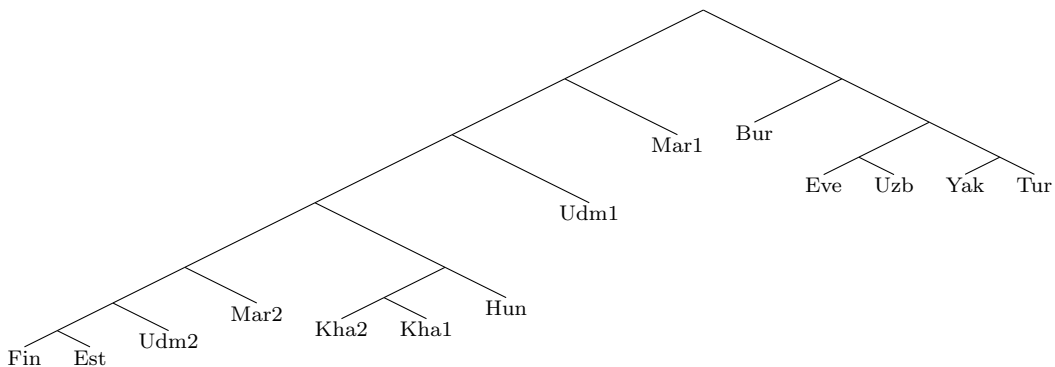

\scriptsize{
$$\Tree[ [  [  [  [  [  [ Fin Est ]  Udm2 ]  Mar2 ]   [  [ Kha2 Kha1 ]  Hun ]  ]  Udm1 ]  Mar1 ] [ Bur [ [ Eve Uzb ] [ Yak Tur ]  ] ]   ]$$
}
\caption{\logdetnj construction:  Finnish (Fin), Estonian (Est),   Udmurt-2 (Udm2), Mari-2, (Ma2), Khanty-2 (Kha2), Khanty-1 (Kha1),   Hungarian (Hun), Udmurt-1 (Udm1)   Mari-1 (Mar1),  Buryat (Bur), Eve,  Uzbek (Uzb), Yakut (Yak), Turkish, (Tur).}
\end{figure}
 
\begin{figure}[H]
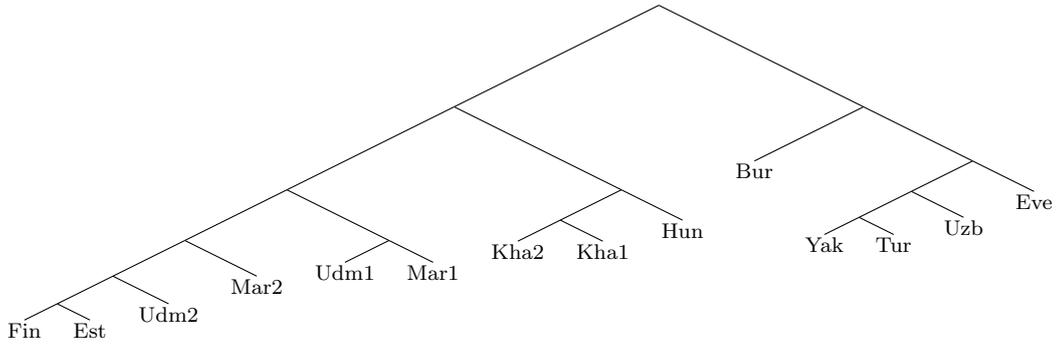

\scriptsize{
$$
\Tree[ 
[  [  [  [  [ Fin Est ]  Udm2 ]  Mar2 ]   [ Udm1 Mar1 ]  ] [  [ Kha2 Kha1 ]  Hun ]  ]
[ Bur    [ [  [ Yak Tur ]  Uzb ]   Eve ] ] ]
$$
}
\caption{Modified Jaccard construction for Uralic and Altaic languages.}
\end{figure}
 
\subsection{The full Indo-European family}\label{full_ie_family}
 
In the analysis of the full Indo-European tree we see that the modified Jaccard index slightly outperforms the $\logdetnj$ construction. Both methods misplace Welsh. Note that Irish and Welsh were degenerate in this dataset (namely they have the same values of all the recorded parameters), so we retained only Welsh. Ceolin et al kept both and so obtained that Welsh and Irish place together. The $\logdetnj$ tree now loses the West and North Germanic split, while the modified Jaccard index is more stable.

\begin{figure}[H]
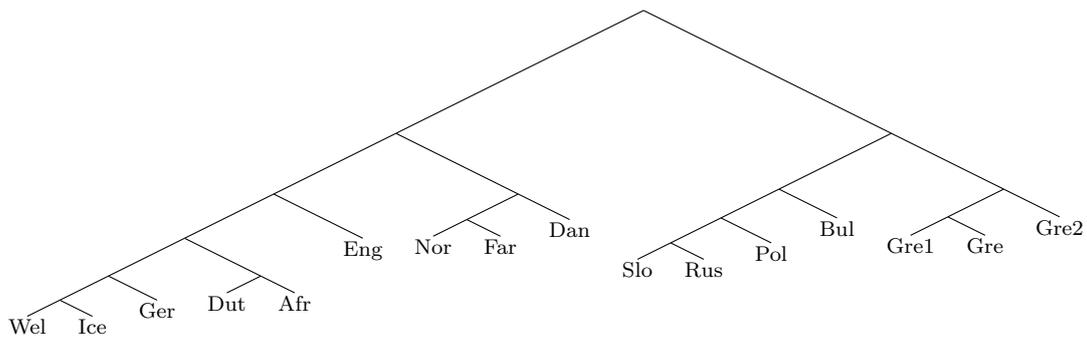

{\scriptsize
$$\Tree[ [  [  [  [  [ Wel Ice ]  Ger ]   [ Dut Afr ]  ]  Eng ]   [  [ Nor Far ]  Dan ]  ]      [  [  [  [ Slo Rus ]  Pol ]  Bul ]   [  [ Gre1 Gre ]  Gre2 ]  ]   ]$$
}
\caption{\logdetnj:Germanic, Slavic and Greek languages from the full Indo-European languages set:Dutch (Dut), Afrikaans (Afr), German (Ger), English (Eng), Faroese (Far), Danish (Dan), Norwegian (Nor), Welsh (Wel), Icelandic (Ice), Slovenian (Slo), Russian (Rus), Polish (Pol), Bulgarian (Bul), Greek-Calabria
N.1 (Gre1), Greek (Gre), Greek-Calabria N.2 (Gre2)}
\end{figure}
 
\begin{figure}[H]
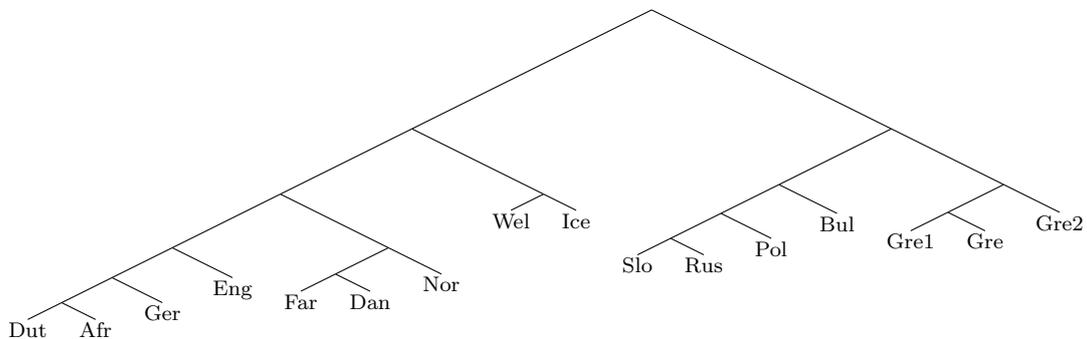

{\scriptsize
$$\Tree[ [  [  [  [  [ Dut Afr ]  Ger ]  Eng ]   [  [ Far Dan ]  Nor ]  ]   [ Wel Ice ]  ] [     [  [  [ Slo Rus ]  Pol ]  Bul ]     [  [ Gre1 Gre ]  Gre2 ]  ]  ]$$
}
\caption{Modified Jaccard:Germanic, Slavic and Greek languages from the full Indo-European languages set}
\end{figure}
 
The Romance subtrees now have correct large-scale structure with Italian dialects forming their own group. The Italian-Portuguese-Spanish conflation that we saw in SSWL-LanGeLin analysis becomes clearer given how these three separate out. 
 
\begin{figure}[H]
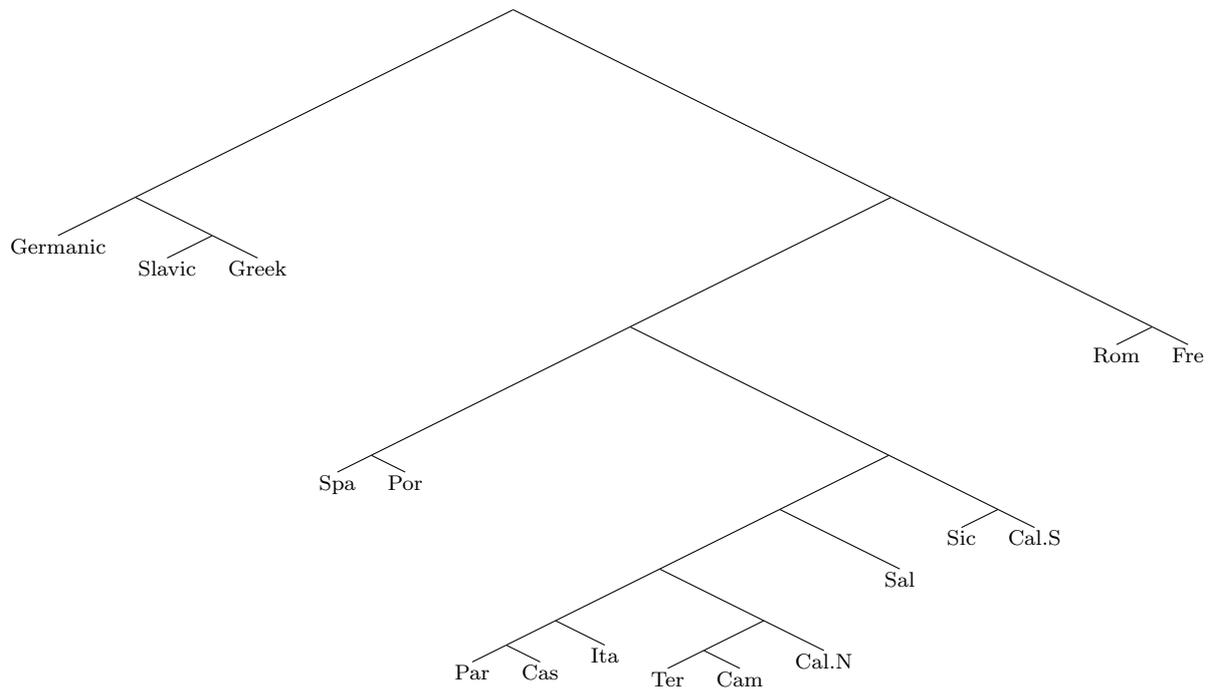

{\scriptsize
$$\Tree[
[  Germanic   [  Slavic   Greek  ]  ]
[     [ [ Spa Por ] [  [  [  [  [ Par Cas ]  Ita ]   [  [ Ter Cam ]  Cal.N ]  ]  Sal ]   [ Sic Cal.S ]  ]     ]  
[ Rom Fre ] ]
]
$$
}
\caption{\logdetnj: Germanic, Slavic and Hellenic languages from the full Indo-European languages set. Note the tree lengths are not to scale; we are considering tree topology only.}
\end{figure}
 
We see that modified Jaccard and $\logdetnj$ both correctly recover that the Indo-Iranian languages split off from the European ones, unlike \namecite{longo2020} where they are mixed between the Greek and Romance subtrees. 
 
\begin{figure}[H]
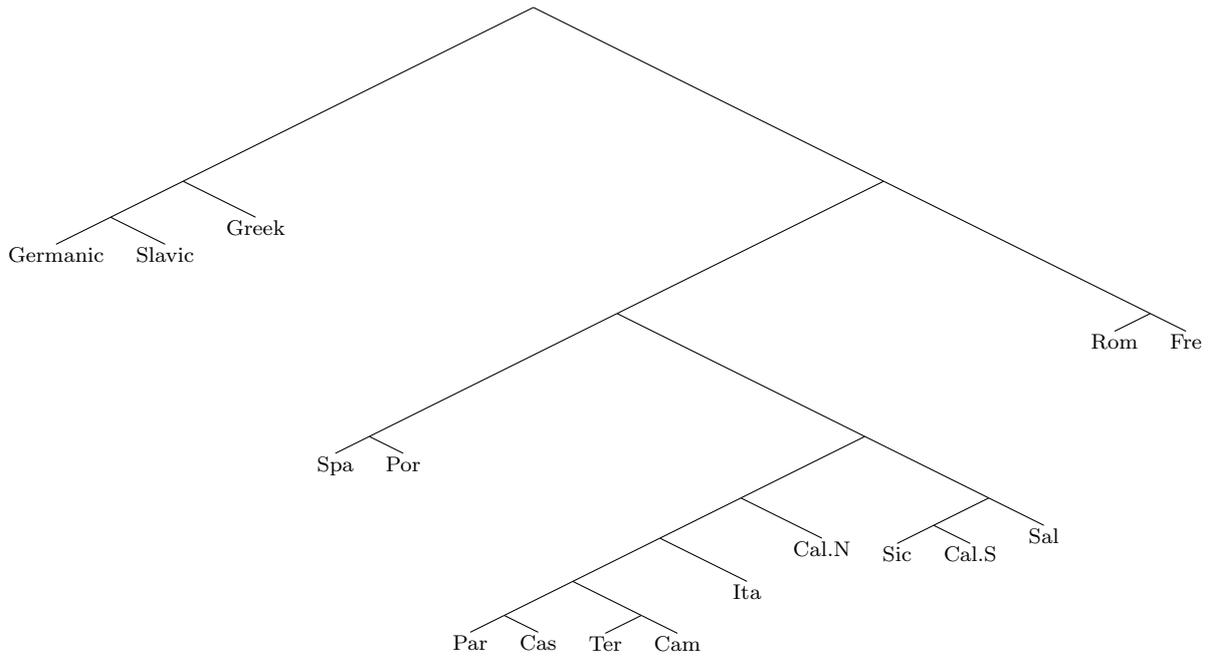

{\scriptsize
$$\Tree[ [  [  Germanic   Slavic  ]   Greek ] [    [ [ Spa Por ]       [  [  [  [  [ Par Cas ]   [ Ter Cam ]  ]  Ita ]  Cal.N ]   [  [ Sic Cal.S ]  Sal ]  ]     ]    [ Rom   Fre ] ] ] $$
}
\caption{Modified Jaccard: Romance languages from the full Indo-European languages set; the Italian dialects are now correctly forming their own subtree.}
\end{figure}
 
\begin{figure}[H]
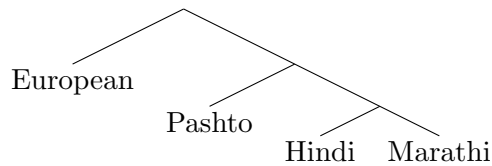

$$\Tree [  European [ Pashto  [ Hindi   Marathi ] ] ] $$
\caption{Both modified Jaccard and $\logdet$ place the Indo-Iranian languages identically in relation to the European languages}
\end{figure}
 
Putting this together we note that both the $\logdetnj$ and modified Jaccard index/infinite site model recover the large-scale structure in Indo-European languages. Modified Jaccard index with neighbor joining is more stable than logdet -- the asymmetry in how syntactic structure change across language families (and possibly how they evolved) is hardwired into the modified Jaccard metric, while lodget/general Markov model may overfit, because of its flexibility. These two methods of reconstruction, reflecting different evolutionary models, are mostly in agreement, disagreeing on finer scale structures where effects of deviations from the assumptions of the underlying models affect each differently and to varying significance.  
 
The following table (table~\ref{rf_values}) quantifies the observed differences between modified Jaccard and logdet based trees using the Robinson--Fould metric. 
 
\begin{table}[H]
\centering
\begin{tabular}{lrr}
\toprule
Family &  rf &  rf\_normalized \\
\midrule
Indo-Iranian &   2 &       0.500000 \\
Germanic &   4 &       0.400000 \\
Balto-Finnic, Ugric and Altaic &   6 &       0.272727 \\
Indo-European &   8 &       0.307692 \\
Germanic, Slavic and Greek\\ languages from the full Indo-European languages set &   8 &       0.307692 \\
\bottomrule
\end{tabular}\caption{Robinson-Fould's distance (rf) and normalized Robinson-Fould's distance (rf\_normalized)  between the modified Jaccard and logdet trees. \label{rf_values}}
\end{table}
 
\section{Conclusion}
 
This analysis was aimed towards trying to understand how well the general Markov and approximate infinite site model describe the syntactic structures data to get insight into how human languages change. The point has not been to derive a metric, possibly abstract, that yields the expected phylogenetic relationships but to understand how well the phylogenetic relationship can be modelled by a type of processes we do understand well. This is an important question, since if we know the process is well described by a Markov model, we can have much more confidence in introducing complexity and fine tuning the models, whereas in cases where we do not know if alternatives to Markov model are equally applicable, the danger of fitting an incorrect model using an abundance of parameters becomes real. 
 
We summarize in Tables~\ref{summary_table} and \ref{summary_table2} some of the comparative analysis and we make a note of how well the Markov model recovers the expected linguistics relationships. We can therefore gain an insight as to when the Markov model assumption can be safely applied. Then, using modified Jaccard metric derived alternative, we recap when the Markov model becomes ill suited.   
 
\begin{table}[H]
\begin{tabular}{p{3cm}|p{3cm}|p{6cm}}
Tree I & Tree II & Results/Insight  \\
\hline
Gemanic family: \logdetnj tree & UPGMA trees with $\ell_p$ metrics, persistent connected components of \S 6.4 of \namecite{marcolli_topling}& \logdetnj tree recovers the North/West Germanic split while others fail. This can be interpreted as support for a Markov model.  \\
\hline
Slavic family: \logdetnj tree & \namecite{Longo2013,SOBM,nurbakova} & There is disagreement on the position of Slovenian; however, the character data does not separate Slovenian from Russian indicating close syntactic proximity. The Markov model is equally well supported as the alternatives.  \\
\hline
North Eurasian languages: \logdetnj  & Topological method of \namecite{marcolli_topling} & \logdetnj is in better agreement with known linguistic relationships and correctly recovers the large-scale structure of the family\\
\hline
Early Indo-European languages: \logdetnj tree & Trees from \namecite{bouckaert, RWT,rexova2003cladistic} & \logdetnj tree, $T_{\text{orig}}$, agrees with \namecite{bouckaert} while $T_4$ does not. The sensitive dependence on a small number of parameters and disagreement with phylogenetic invariants suggests that the parameters are not well described by a Markov model, but the phylogenetic invariants derived from the Markov model still select the correct \cite{RWT} tree.\\ 
\hline
Romance languages: \logdetnj tree & Historically accepted trees, tree from \namecite{SAHWM} & Misplacement similar to what \namecite{SAHWM} observed; exploration of parameter values reveals very few parameters separate languages in this family. The phylogenetic invariants select \logdetnj tree over historically correct tree suggesting that general Markov model does not describe this family well, likely dues to high degree of relatedness (cf. section \ref{fig_romance_param_clusters}).  
 
The syntactic parameters here are likely not well described by a Markov model.\\ 
\hline
\end{tabular}\caption{Table summarizing the comparative analysis against phylogenetic trees from linguistics literature. Syntactic parameters for languages that are too closely related are not well described by \logdetnj/general Markov models. \label{summary_table}} 
\end{table}
 
\begin{table}[H]
\begin{tabular}{p{4cm}|p{8cm}}
Family & \logdetnj vs. modified Jaccard results  \\
\hline
Indo-Iranian+Dravidian & Modified Jaccard correctly separates out the two Dravidian languages while \logdetnj fails. \\
\hline
Greco-Romance & Both metrics produce identical trees.\\
\hline
Germanic & Modified Jaccard places Danish with Faroese (versus expected Norwegian) while \logdetnj fails to place Dutch with Afrikaans, and also mixes Faroese-Norwegian-Danish. Ceolin et al's BEAST construction also places German with Afrikaan. The close linguistic relationships are not well modelled by the Markov model. \\
\hline
Balto-Finnic, Ugric and Altaic & Modified Jaccard outperforms \logdetnj; misplacements by \logdetnj are similar to those from Ceolin et al's BEAST tree. \\ 
\hline
Indo-European & Modified Jaccard slightly ourperforms \logdetnj; the subtree of Germanic languages is the same for modified Jaccard, but not so for \logdetnj; both correctly recover the  splitting off of Indo-Iranian languages from Indo-European. The correct large scale structure within the Romance family is also recovered by both.\\
\hline
\end{tabular}\caption{Summarizing the comparative analysis between \logdetnj trees and trees using neighbor joining with the modified Jaccard metric of \namecite{ceolin2021boundaries} which can be interpreted as describing an infinite site evolutionary model: the modified Jaccard outperforms the Markov model-based approach for closely related languages. We hypothesize that the ability to mutate a site more than once afforded by the Markov model is not useful, making it less suited than modified Jaccard. \label{summary_table2}}
\end{table}
 
There are two primary difficulties that arise in trying to evaluate how well the syntactic parameters data fits the Markov model.  Both can be interpreted as being indicative of how processes underlying the data are deviating from the Markov model. 
 
\begin{itemize}
\item Linguistic relationships across multiple families are often not stably reconstructed. 
\item Languages with high degree of relatedness are difficult to place. 
\end{itemize}
 
The first can be addressed by noting that the hypothesis of a single root from which they can be considered to have evolved may be accurate for linguistic subfamilies, but ancestral languages and proto-languages lying behind sufficiently different linguistic families are highly hypothetical (the contested Ural-Altaic hypothesis being one such example), hence simply trying to fit diverse syntactic data across a broad range of language families to such a model with a single root should not be expected to be very meaningful. 
 
The second suggests that the tree topology is insufficient to capture how they have influenced each other, and the small set of syntactic parameters that separate them make the reconstruction statistically not very robust. There's also a type of sampling problem present: the representatives of language families are not generated as random samples from the process acting on the family; with a single close relationship in a family, coupled with the small size of the families, the biases become extremely significant. An algorithm, like neighbor joining, that uses both local information (the pairwise distances) and global information (pairwise distances to the rest of the tree), is likely to be thrown off by these biases. This is evident from the example where, while Spanish and Portuguese share the highest similarity in terms of the logdet metric, their placement does not reflect this. This is also supported by the observation that larger sets from within the same family tend to give a more correct picture of the phylogenetic relationships, even though there is a tradeoff that the larger collection may be less likely to be described by a simple model.   
 
This leads us to conclude that phylogenetic inference at larger scales, across many families, using syntactic structures data alone is unlikely to be better than a crude approximation of the underlying truth. Based on our results, the phylogenetic signal from syntactic structures is only reliable when the languages being considered belong to the same family -- the assumption that they come from a single root is reasonable, but at the same time are unlikely to have so much interaction that the tree topology becomes inadequate -- for example: Portuguese and Spanish in the romance family. 
 
We believe that more significant theoretical work is needed on dynamical models of language change at the syntactic level, that can replace the Markov hypothesis with a more accurate model, tailored to linguistic needs, that can be used for better phylogenetic inference of relevance to
historical linguistics. \\
 
\emph{Acknowledgement.} We would like to thank Andrea Ceolin for thoughtful feedback on the previous version of this paper that motivated this revision. 
The second author is partially supported by NSF grant DMS-2104330. 

\emph{Code/Data availability.} The code and data used in this paper are available at \url{https://github.com/minorllama/syntactic_structures_phylogenetics}

\newpage
 
\bibliography{references}

\begin{thebibliography}{60}
\providecommand{\natexlab}[1]{#1}
\providecommand{\url}[1]{\texttt{#1}}
\expandafter\ifx\csname urlstyle\endcsname\relax
  \providecommand{\doi}[1]{doi: #1}\else
  \providecommand{\doi}{doi: \begingroup \urlstyle{rm}\Url}\fi

\bibitem[Allman and Rhodes(2008)]{AllRho}
E.~Allman and J.~Rhodes.
\newblock Phylogenetic ideals and varieties for general markov models.
\newblock \emph{Adv. Appl. Math.}, 40:\penalty0 127--148, 2008.

\bibitem[Allman et~al.(2012)Allman, Rhodes, and Sullivant]{allman2012mixture}
Elizabeth~S Allman, John~A Rhodes, and Seth Sullivant.
\newblock When do phylogenetic mixture models mimic other phylogenetic models?
\newblock \emph{Systematic biology}, 61\penalty0 (6):\penalty0 1049--1059,
  2012.

\bibitem[Baker(2002)]{Baker}
Mark~C. Baker.
\newblock \emph{The Atoms of Language}.
\newblock Basic Books, 2002.

\bibitem[Biberauer(2008)]{Biberauer}
Theresa Biberauer.
\newblock \emph{The Limits of Syntactic Variation}.
\newblock John Benjamins Publishing, 2008.

\bibitem[Bouckaert et~al.(2012)Bouckaert, Lemey, Dunn, Greenhill, Alekseyenko,
  Drummond, Gray, Suchard, and Atkinson]{bouckaert}
Remco Bouckaert, Philippe Lemey, Michael Dunn, Simon~J Greenhill, Alexander~V
  Alekseyenko, Alexei~J Drummond, Russell~D Gray, Marc~A Suchard, and Quentin~D
  Atkinson.
\newblock Mapping the origins and expansion of the indo-european language
  family.
\newblock \emph{Science}, 337\penalty0 (6097):\penalty0 957--960, 2012.

\bibitem[Bouckaert et~al.(2014)Bouckaert, Heled, K{\"u}hnert, Vaughan, Wu, Xie,
  Suchard, Rambaut, and Drummond]{beast2014}
Remco Bouckaert, Joseph Heled, Denise K{\"u}hnert, Tim Vaughan, Chieh-Hsi Wu,
  Dong Xie, Marc~A Suchard, Andrew Rambaut, and Alexei~J Drummond.
\newblock Beast 2: a software platform for bayesian evolutionary analysis.
\newblock \emph{PLoS Comput Biol}, 10\penalty0 (4):\penalty0 e1003537, 2014.

\bibitem[Ceolin et~al.(2020)Ceolin, Guardiano, Irimia, and
  Longobardi]{longo2020}
Andrea Ceolin, Cristina Guardiano, Monica~Alexandrina Irimia, and Giuseppe
  Longobardi.
\newblock Formal syntax and deep history.
\newblock \emph{Frontiers in psychology}, 11:\penalty0 2384, 2020.

\bibitem[Ceolin et~al.(2021)Ceolin, Guardiano, Longobardi, Irimia, Bortolussi,
  and Sgarro]{ceolin2021boundaries}
Andrea Ceolin, Cristina Guardiano, Giuseppe Longobardi, Monica~Alexandrina
  Irimia, Luca Bortolussi, and Andrea Sgarro.
\newblock At the boundaries of syntactic prehistory.
\newblock \emph{Philosophical Transactions of the Royal Society B},
  376\penalty0 (1824):\penalty0 20200197, 2021.

\bibitem[Chomsky(1981)]{chomsky_govt}
Noam Chomsky.
\newblock \emph{Lectures on Government and Binding}.
\newblock Walter de Gruyter, 1981.

\bibitem[Chomsky and Lasnik(1993)]{ChoLa}
Noam Chomsky and Howard Lasnik.
\newblock The theory of principles and parameters.
\newblock In \emph{Syntax: An international handbook of contemporary research},
  pages 506--569. Walter de Gruyter, 1993.

\bibitem[Collins(2010)]{collins_sswl}
Chris Collins.
\newblock Syntactic structures of the world's language: A cross-linguistic
  database.
\newblock 2010.
\newblock 27 September 2010, Colloquium:
  \url{https://ling.yale.edu/syntactic-structures-worlds-language-cross-linguistic-database}.

\bibitem[Dryer and Haspelmath(2013)]{wals}
Matthew~S. Dryer and Martin Haspelmath, editors.
\newblock \emph{WALS Online}.
\newblock Max Planck Institute for Evolutionary Anthropology, Leipzig, 2013.
\newblock URL \url{https://wals.info/}.

\bibitem[Durrett(2019)]{durrett2019probability}
Rick Durrett.
\newblock \emph{Probability: theory and examples}, volume~49.
\newblock Cambridge university press, 2019.

\bibitem[Eriksson(2006)]{eriksson_thesis}
Nicholas~Karl Eriksson.
\newblock \emph{Algebraic combinatorics for computational biology}.
\newblock PhD thesis, University of California, Berkeley, 2006.

\bibitem[Felsenstein(2004)]{fels2004}
Joseph Felsenstein.
\newblock \emph{Inferring phylogenies}, volume~2.
\newblock Sinauer associates Sunderland, MA, 2004.

\bibitem[Gascuel and Steel(2006)]{nj_revealed}
Olivier Gascuel and Mike Steel.
\newblock Neighbor-joining revealed.
\newblock \emph{Molecular biology and evolution}, 23\penalty0 (11):\penalty0
  1997--2000, 2006.

\bibitem[Gray et~al.(2009)Gray, Drummond, and Greenhill]{greenhill_pacific}
Russell~D Gray, Alexei~J Drummond, and Simon~J Greenhill.
\newblock Language phylogenies reveal expansion pulses and pauses in pacific
  settlement.
\newblock \emph{science}, 323\penalty0 (5913):\penalty0 479--483, 2009.

\bibitem[Guardiano et~al.(2016)Guardiano, Michelioudakis, Ceolin, Irimia,
  Longobardi, Radkevich, Sitaridou, and Silvestri]{microvar}
C.~Guardiano, D.~Michelioudakis, A.~Ceolin, M.~Irimia, G.~Longobardi,
  N.~Radkevich, I.~Sitaridou, and G.~Silvestri.
\newblock South by southeast. a syntactic approach to greek and romance
  microvariation.
\newblock \emph{L'Italia Dialettale}, 77:\penalty0 95--166, 2016.

\bibitem[Hoffmann et~al.(2021)Hoffmann, Bouckaert, Greenhill, and
  K{\"u}hnert]{Greenhill}
Konstantin Hoffmann, Remco Bouckaert, Simon~J Greenhill, and Denise
  K{\"u}hnert.
\newblock Bayesian phylogenetic analysis of linguistic data using beast.
\newblock \emph{Journal of Language Evolution}, pages 1--17, 09 2021.

\bibitem[Karimi and Piattelli-Palmarini(2017)]{LingAn}
S.~Karimi and M.~Piattelli-Palmarini.
\newblock Special issue on parameters.
\newblock \emph{Linguistic Analysis}, 41\penalty0 (3-4), 2017.

\bibitem[Kazakov et~al.(2018)Kazakov, Cordoni, Algahtani, Ceolin, Irimia, Kim,
  Michelioudakis, Radkevich, Guardiano, and Longobardi]{evolang12}
Dimitar Kazakov, Guido Cordoni, Eyad Algahtani, Andrea Ceolin, Monica~A.
  Irimia, Shin-Sook Kim, Dimitris Michelioudakis, Nina Radkevich, Cristina
  Guardiano, and Giuseppe Longobardi.
\newblock Learning implicational models of universal grammar parameters.
\newblock In C.~Cuskley, M.~Flaherty, H.~Little, Luke McCrohon, A.~Ravignani,
  and T.~Verhoef, editors, \emph{The Evolution of Language: Proceedings of the
  12th International Conference (EVOLANGXII)}. NCU Press, 2018.
\newblock \doi{10.12775/3991-1.048}.
\newblock URL
  \url{http://evolang.org/torun/proceedings/papertemplate.html?p=176}.

\bibitem[Koopman(2011)]{koopman_sswl}
Hilda Koopman.
\newblock Sswl syntactic structures of the world's languages: An open-ended
  database for the linguistic community and by the linguistic community
  \url{http://sswl.railsplayground.net/} mit 50, 12 2011.

\bibitem[Lake(1994)]{lake1994}
James~A Lake.
\newblock Reconstructing evolutionary trees from dna and protein sequences:
  paralinear distances.
\newblock \emph{Proceedings of the National Academy of Sciences}, 91\penalty0
  (4):\penalty0 1455--1459, 1994.

\bibitem[Longobardi(2012{\natexlab{a}})]{Lo12}
Giuseppe Longobardi.
\newblock Convergence in parametric phylogenies. homoplasy or principled
  explanation?
\newblock In Charlotte Galves, Sonia Cyrino, Ruth Lopes, Filomena Sandalo, and
  Juanito Avelar, editors, \emph{Parameter Theory and Linguistic Change}.
  Oxford University Press, 2012{\natexlab{a}}.
\newblock \doi{10.1093/acprof:oso/9780199659203.001.0001}.

\bibitem[Longobardi(2012{\natexlab{b}})]{homoplasy}
Giuseppe Longobardi.
\newblock Convergence in parametric phylogenies: Homoplasy or principled
  explanation?
\newblock In \emph{Parameter Theory and Language Change}, pages 304--319.
  Oxford University Press, 2012{\natexlab{b}}.

\bibitem[Longobardi(2017{\natexlab{a}})]{LongoNew}
Giuseppe Longobardi.
\newblock Principles, parameters, and schemata: a constructivist {UG}.
\newblock \emph{Linguistic Analysis}, 41\penalty0 (3-4):\penalty0 517--556,
  2017{\natexlab{a}}.

\bibitem[Longobardi(2017{\natexlab{b}})]{langelin}
Giuseppe Longobardi.
\newblock Principles, parameters, and schemata. a constructivist ug.
\newblock \emph{Linguistic Analysis}, \penalty0 (41):\penalty0 517--557,
  2017{\natexlab{b}}.
\newblock
  \url{https://www.york.ac.uk/language/research/projects/langelin/#tab-1}.

\bibitem[Longobardi and Guardiano(2009)]{LongGua}
Giuseppe Longobardi and Cristina Guardiano.
\newblock Evidence for syntax as a signal of historical relatedness.
\newblock \emph{Lingua}, 119:\penalty0 1679--1706, 2009.

\bibitem[Longobardi et~al.(2013)Longobardi, Guardiano, Silvestri, Boattini, and
  Ceolin]{Longo2013}
Giuseppe Longobardi, Cristina Guardiano, Giuseppina Silvestri, Alessio
  Boattini, and Andrea Ceolin.
\newblock Toward a syntactic phylogeny of modern indo-european languages.
\newblock \emph{Journal of Historical Linguistics}, 3\penalty0 (1):\penalty0
  122--152, 2013.

\bibitem[Ma et~al.(2008)Ma, Ratan, Raney, Suh, Miller, and Haussler]{ma_ism}
Jian Ma, Aakrosh Ratan, Brian~J. Raney, Bernard~B. Suh, Webb Miller, and David
  Haussler.
\newblock The infinite sites model of genome evolution.
\newblock \emph{Proceedings of the National Academy of Sciences}, 105\penalty0
  (38):\penalty0 14254--14261, 2008.

\bibitem[Marcolli(2016)]{marcolli_ling_ecc}
Matilde Marcolli.
\newblock Syntactic parameters and a coding theory perspective on entropy and
  complexity of language families.
\newblock \emph{Entropy}, 18\penalty0 (4):\penalty0 Paper No. 110, 17, 2016.
\newblock ISSN 1099-4300.
\newblock \doi{10.3390/e18040110}.
\newblock URL
  \url{https://doi-org.clsproxy.library.caltech.edu/10.3390/e18040110}.

\bibitem[Matsen and Steel(2007)]{matsen_mimicry}
Frederick~A Matsen and Mike Steel.
\newblock Phylogenetic mixtures on a single tree can mimic a tree of another
  topology.
\newblock \emph{Systematic Biology}, 56\penalty0 (5):\penalty0 767--775, 2007.

\bibitem[Murawaki(2018)]{murawaki}
Yugo Murawaki.
\newblock Analyzing correlated evolution of multiple features using latent
  representations.
\newblock In \emph{Proceedings of the 2018 Conference on Empirical Methods in
  Natural Language Processing}, pages 4371--4382, 2018.

\bibitem[Nicholls and Gray(2008)]{Nicholls-Gray-2008}
Geoff~K Nicholls and Russell~D Gray.
\newblock Dated ancestral trees from binary trait data and their application to
  the diversification of languages.
\newblock \emph{Journal of the Royal Statistical Society: Series B (Statistical
  Methodology)}, 70:\penalty0 545--566, 2008.

\bibitem[Niyogi(2006)]{niyogi}
Partha Niyogi.
\newblock \emph{The computational nature of language learning and evolution},
  volume~43 of \emph{Current Studies in Linguistics}.
\newblock MIT Press, Cambridge, MA, 2006.
\newblock ISBN 0-262-14094-2.

\bibitem[Niyogi and Berwick(1997)]{berwickniyogi97}
Partha Niyogi and Robert~C. Berwick.
\newblock A dynamical systems model for language change.
\newblock \emph{Complex Systems}, pages 161--204, 1997.

\bibitem[Nurbakova et~al.(2013)Nurbakova, Rusakov, and Alexandrov]{nurbakova}
Diana Nurbakova, Sergey Rusakov, and Vassil Alexandrov.
\newblock Quantifying uncertainty in phylogenetic studies of the slavonic
  languages.
\newblock \emph{Procedia Computer Science}, 18:\penalty0 2269--2277, 2013.

\bibitem[O'Donnell(2014)]{odonnell_boolean}
Ryan O'Donnell.
\newblock \emph{Analysis of boolean functions}.
\newblock Cambridge University Press, 2014.

\bibitem[Ortegaray et~al.(2018)Ortegaray, Berwick, and Marcolli]{marcolli_hk}
Andrew Ortegaray, Robert~C. Berwick, and Matilde Marcolli.
\newblock Heat kernel analysis of syntactic structures.
\newblock \emph{CoRR}, abs/1803.09832, 2018.
\newblock URL \url{http://arxiv.org/abs/1803.09832}.

\bibitem[Pachter and Sturmfels(2005)]{pachter_algstats}
Lior Pachter and Bernd Sturmfels.
\newblock \emph{Algebraic statistics for computational biology}, volume~13.
\newblock Cambridge university press, 2005.

\bibitem[Pachter and Sturmfels(2007)]{PaSturm}
Lior Pachter and Bernd Sturmfels.
\newblock The mathematics of phylogenomics.
\newblock \emph{SIAM Review}, 49\penalty0 (1):\penalty0 3--31, 2007.

\bibitem[Pagel and Meade(2004)]{Pagel-Meade}
M.~Pagel and A.~Meade.
\newblock A phylogenetic mixture model for detecting pattern-heterogeneity in
  gene sequence or character-state data.
\newblock \emph{Syst Biol .}, 53\penalty0 (4):\penalty0 571--81, Aug 2004.

\bibitem[Park et~al.(2017)Park, Boettcher, Zhao, Mun, Yuh, Kumar, and
  Marcolli]{kanerva}
J.J. Park, R.~Boettcher, A.~Zhao, A.~Mun, K.~Yuh, V.~Kumar, and M.~Marcolli.
\newblock Prevalence and recoverability of syntactic parameters in sparse
  distributed memories.
\newblock In \emph{Geometric Structures of Information 2017. Lecture Notes in
  Computer Science, Vol.~10589}, pages 1--8. Springer, 2017.

\bibitem[Perelysvaig and Lewis(2015)]{IEcontroversy}
A.~Perelysvaig and M.W. Lewis.
\newblock \emph{The Indo-European controversy: facts and fallacies in
  Historical Linguistics}.
\newblock Cambridge University Press, 2015.

\bibitem[Piispanen(2013)]{Piispanen_2013}
Peter Piispanen.
\newblock The uralic-yukaghiric connection revisited: Sound correspondences of
  geminate clusters.
\newblock \emph{Suomalais-Ugrilaisen Seuran Aikakauskirja}, 2013\penalty0
  (94):\penalty0 165--197, tammi 2013.
\newblock \doi{10.33340/susa.82515}.
\newblock URL \url{https://journal.fi/susa/article/view/82515}.

\bibitem[Port et~al.(2018)Port, Gheorghita, Guth, Clark, Liang, Dasu, and
  Marcolli]{port2018}
Alexander Port, Iulia Gheorghita, Daniel Guth, John~M. Clark, Crystal Liang,
  Shival Dasu, and Matilde Marcolli.
\newblock Persistent topology of syntax.
\newblock \emph{Mathematics in Computer Science}, 12\penalty0 (1):\penalty0
  33--50, Mar 2018.
\newblock ISSN 1661-8289.
\newblock \doi{10.1007/s11786-017-0329-x}.
\newblock URL \url{https://doi.org/10.1007/s11786-017-0329-x}.

\bibitem[Port et~al.(2019)Port, Karidi, and Marcolli]{marcolli_topling}
Alexander Port, Taelin Karidi, and Matilde Marcolli.
\newblock Topological analysis of syntactic structures.
\newblock \emph{CoRR}, abs/1903.05181, 2019.
\newblock URL \url{http://arxiv.org/abs/1903.05181}.

\bibitem[Rexov{\'a} et~al.(2003)Rexov{\'a}, Frynta, and
  Zrzav{\`y}]{rexova2003cladistic}
Kate{\v{r}}ina Rexov{\'a}, Daniel Frynta, and Jan Zrzav{\`y}.
\newblock Cladistic analysis of languages: Indo-european classification based
  on lexicostatistical data.
\newblock \emph{Cladistics}, 19\penalty0 (2):\penalty0 120--127, 2003.

\bibitem[Ringe et~al.(2002)Ringe, Warnow, and Taylor]{RWT}
D.~Ringe, T.~Warnow, and A.~Taylor.
\newblock Indo-european and computational cladistics.
\newblock \emph{Transactions of the Philological Society}, 100:\penalty0
  59--129, 2002.

\bibitem[Rizzi(2017)]{Rizzi}
Luigi Rizzi.
\newblock On the format and locus of parameters: the role of morphosyntactic
  features.
\newblock \emph{Linguistic Analysis}, 41\penalty0 (3-4):\penalty0 159--191,
  2017.

\bibitem[Saitou and Nei(1987)]{saitou1987neighbor}
Naruya Saitou and Masatoshi Nei.
\newblock The neighbor-joining method: a new method for reconstructing
  phylogenetic trees.
\newblock \emph{Molecular biology and evolution}, 4\penalty0 (4):\penalty0
  406--425, 1987.

\bibitem[Semple et~al.(2003)Semple, Steel, et~al.]{semple_phylo}
Charles Semple, Mike Steel, et~al.
\newblock \emph{Phylogenetics}, volume~24.
\newblock Oxford University Press on Demand, 2003.

\bibitem[Shu and Marcolli(2017)]{shu_marcolli}
Kevin Shu and Matilde Marcolli.
\newblock Syntactic structures and code parameters.
\newblock \emph{Math. Comput. Sci.}, 11\penalty0 (1):\penalty0 79--90, 2017.
\newblock ISSN 1661-8270.
\newblock \doi{10.1007/s11786-017-0298-0}.
\newblock URL
  \url{https://doi-org.clsproxy.library.caltech.edu/10.1007/s11786-017-0298-0}.

\bibitem[Shu et~al.(2017)Shu, Ortegaray, Berwick, and Marcolli]{SOBM}
Kevin Shu, Andrew Ortegaray, Robert~C. Berwick, and Matilde Marcolli.
\newblock Phylogenetics of indo-european language families via an
  algebro-geometric analysis of their syntactic structures.
\newblock \emph{CoRR}, abs/1712.01719, 2017.
\newblock URL \url{http://arxiv.org/abs/1712.01719}.

\bibitem[Shu et~al.(2018)Shu, Aziz, Huynh, Warrick, and Marcolli]{SAHWM}
Kevin Shu, Sharjeel Aziz, Vy-Luan Huynh, David Warrick, and Matilde Marcolli.
\newblock Syntactic phylogenetic trees.
\newblock In \emph{Foundations of mathematics and physics one century after
  {H}ilbert}, pages 417--441. Springer, Cham, 2018.

\bibitem[{\v{S}}tefankovi{\v{c}} and
  Vigoda(2007{\natexlab{a}})]{vstefanko_mimicry_a}
Daniel {\v{S}}tefankovi{\v{c}} and Eric Vigoda.
\newblock Phylogeny of mixture models: Robustness of maximum likelihood and
  non-identifiable distributions.
\newblock \emph{Journal of Computational Biology}, 14\penalty0 (2):\penalty0
  156--189, 2007{\natexlab{a}}.

\bibitem[{\v{S}}tefankovi{\v{c}} and
  Vigoda(2007{\natexlab{b}})]{vstefanko_mimicry_b}
Daniel {\v{S}}tefankovi{\v{c}} and Eric Vigoda.
\newblock Pitfalls of heterogeneous processes for phylogenetic reconstruction.
\newblock \emph{Systematic biology}, 56\penalty0 (1):\penalty0 113--124,
  2007{\natexlab{b}}.

\bibitem[Stumpf et~al.(2017)Stumpf, Smith, Lenz, Schuppert, M{\"u}ller, Babtie,
  Chan, Stumpf, Please, Howison, et~al.]{stumpf2017}
Patrick~S Stumpf, Rosanna~CG Smith, Michael Lenz, Andreas Schuppert,
  Franz-Josef M{\"u}ller, Ann Babtie, Thalia~E Chan, Michael~PH Stumpf, Colin~P
  Please, Sam~D Howison, et~al.
\newblock Stem cell differentiation as a non-markov stochastic process.
\newblock \emph{Cell Systems}, 5\penalty0 (3):\penalty0 268--282, 2017.

\bibitem[Warnow(2017)]{warnow}
Tandy Warnow.
\newblock \emph{Computational Phylogenetics}.
\newblock Cambridge University Press, 2017.

\bibitem[Zou et~al.(2011)Zou, Susko, Field, and Roger]{Zou}
Liwen Zou, Edward Susko, Chris Field, and Andrew~J. Roger.
\newblock {The Parameters of the Barry and Hartigan General Markov Model Are
  Statistically NonIdentifiable}.
\newblock \emph{Systematic Biology}, 60\penalty0 (6):\penalty0 872--875, 04
  2011.
\newblock ISSN 1063-5157.
\newblock \doi{10.1093/sysbio/syr034}.
\newblock URL \url{https://doi.org/10.1093/sysbio/syr034}.

\end{thebibliography}

\end{document}